\documentclass{article}

\usepackage{arxiv}

\usepackage[utf8]{inputenc} 
\usepackage[T1]{fontenc}    
\usepackage{hyperref}       
\usepackage{url}            
\usepackage{booktabs}       
\usepackage{amsfonts}       
\usepackage{nicefrac}       
\usepackage{microtype}      
\usepackage{lipsum}         
\usepackage{graphicx}
\usepackage{natbib}
\usepackage{doi}

\usepackage{array}
\usepackage{multirow}
\usepackage{pifont}
\usepackage{booktabs}
\usepackage{amssymb}
\usepackage{amsmath}
\usepackage{subcaption}
\usepackage{xcolor}
\usepackage{booktabs}
\usepackage{graphicx}
\usepackage{array}
\usepackage[table]{xcolor}
\usepackage{pifont} 
\usepackage{enumitem}
\usepackage{wrapfig}
\usepackage{etoc}

\usepackage{cleveref}       

\makeatletter

\newcommand{\Rmnum}[1]{\expandafter\@slowromancap\romannumeral #1@}
\makeatother

\title{DNA: A Double-Helix Hierarchical Deep Interaction Network Architecture for Industrial Digital Twin}

\date{}

\usepackage{authblk}

\author[1,\dag]{Haoran Yang}
\author[1]{Yanan Xu}
\author[1,*]{Yinan Zhang}
\author[2]{Wenjie Zhang}
\author[1]{Peiyu Liu}
\author[1]{Jiahui Xiang}
\author[1,*]{Wenhai Wang}

\affil[1]{Department of Control Science and Engineering, Zhejiang University, Hangzhou, China}
\affil[2]{Risk Department, Binance}

\renewcommand{\headeright}{}
\renewcommand{\undertitle}{}

\hypersetup{
pdftitle={DNA: A Double-Helix Hierarchical Deep Interaction Network Architecture for Industrial Digital Twin},
pdfsubject={cs.LG},
pdfauthor={Haoran Yang, Yanan Xu, Yinan Zhang, Wenjie Zhang, Peiyu Liu, Jiahui Xiang, Wenhai Wang},
pdfkeywords={Deep learning, Industrial digital twin, Industrial modeling, Data-driven modeling, Three-phase pipeline, Chemical engineering}, 
}

\begin{document}
\maketitle

\begingroup
\renewcommand{\thefootnote}{\fnsymbol{footnote}}
\footnotetext[2]{First author. Email: \texttt{22332111@zju.edu.cn}}  
\footnotetext[1]{Corresponding authors.}                          
\endgroup
\setcounter{footnote}{0}

\vspace{-5mm}

\begin{abstract}
	High-fidelity modeling is fundamental to industrial digital twin, yet a trade-off persists between the physical transparency of mechanistic models and the superior predictive power of data-driven approaches. Existing hybrid methods often fail to bridge this gap, suffering from shallow integration and rigid one-shot modeling strategies that struggle with complex industrial systems. To address these issues, we propose a \textbf{Double-Helix Hierarchical Deep Interaction Network Architecture (DNA)}. Inspired by the double-helix structure of biological \textit{DNA}, our architecture intertwines mechanistic and data-driven modeling strands through two novel interaction strategies, forming a symbiotic architecture in which the two strands mutually enhance predictive capability. Furthermore, the hierarchical structure of DNA allows it to model complex systems in a multi-stage manner through two tailored deep learning frameworks. We evaluate the proposed approach on two distinct industrial systems, where DNA achieves reductions in MAE of 3.704\% and 3.175\%, respectively, compared to the strongest baseline. These results demonstrate the effectiveness of DNA and suggest its potential to provide new insights for the development of industrial digital twins.
\end{abstract}

\section{Introduction}

In industrial digital twin applications, developing high-fidelity models is essential for enabling real-time monitoring, prediction, and optimization \cite{liu2021review,zhang2024promoting}. The fidelity of these models dictates the digital twin system's ability to mirror the intricate physical, chemical, and behavioral dynamics of real-world systems \cite{peterson2025digital}. In industrial practice, mechanistic models are often favored due to their reliability, stability, and physical transparency \cite{schweidtmann2024review}, particularly in scenarios requiring stringent safety standards. Despite their theoretical rigor, mechanistic models suffer from several significant limitations. For instance, they often rely on simplifying assumptions during the modeling process, such as neglecting secondary reactions or assuming perfect mixing, which can introduce systematic biases into their predictions \cite{himmelblau2012basic}. Their predictive performance may also degrade markedly under unfamiliar operating conditions or previously unseen process states \cite{bishnu2023computational}. Morever, mechanistic models typically require substantial modeling effort and computational resources \cite{elsheikh2023control}. These factors further limit the the flexibility of mechanistic approaches in practical applications.

Recently, many researchers have increasingly turned to develop data-driven models, aiming to mitigate the limitations associated with the mechanistic models \cite{ameri2025predictive,mansouri2025models}. Extensive studies demonstrate that data-driven models based on these methods often outperform traditional mechanistic models in terms of prediction accuracy, computational efficiency, and reduced development cost \cite{mane2024digital,pal2025digital,yang2025ultramodel}. However, such data-driven models typically fail to consider the first principles governing the modeled system, making it difficult to generate physically grounded predictions \cite{bradley2022perspectives}. Subsequently, researchers have begun to explore hybrid modeling approaches that integrate mechanistic and data-driven models, aiming to combine the physical transparency of the former with the predictive power of the latter. One of the most representative approaches is structured hybrid modeling (SHM), which explicitly couples the two components through serial, parallel, or combined architectures, forming a clearly structured hybrid system \cite{schweidtmann2024review}, as illustrated in Fig.\ref{fig4-}. However, such methods essentially amount to a simple stacking of mechanistic and data-driven components, following a unidirectional information flow, and thus fail to achieve effective bidirectional interaction between them. Moreover, most existing hybrid modeling approaches adopt a one-shot modeling strategy that attempts to directly predict the final output from the original input in a single step, regardless of the underlying complexity of the target system \cite{su2019architecture,li2020novel}, which, although enabling end-to-end prediction, their performance tends to degrade when applied to large-scale or highly complex industrial systems.

Most importantly, in real-world industrial applications, although data-driven models offer appealing advantages, their black-box nature and limited physical transparence have prevented their widespread adoption in safety-critical industrial settings. Consequently, data-driven modeling approaches remain largely confined to academic research. In contrast, mechanistic models, despite lacking these advantages, are fully interpretable and highly stable, and thus remain the preferred choice in industrial practice. Therefore, enhancing the performance of mechanistic models is of significant industrial value. This naturally leads to a key question: Can we design a modeling architecture that integrates mechanistic and data-driven models in a novel way, allowing it to leverage the superior predictive capability of data-driven models while providing effective feedback to improve the mechanistic models themselves? Such an architecture would allow mechanistic models to benefit from data-driven learning, providing a practical solution (the enhanced mechanistic model) for industrial scenarios where deploying data-driven models is impractical or undesirable.

The answer is yes. Inspired by the double-helix structure of biological \textit{DNA} \cite{klug2004discovery}, we propose a novel Double-Helix Hierarchical \textbf{D}eep Interaction \textbf{N}etwork \textbf{A}rchitecture (DNA) for industrial digital twin. In the proposed architecture, the sequentially connected chains of mechanistic and data-driven models mirror the two long deoxyribonucleotide backbone strands of \textit{DNA}; the interaction between mechanistic models and data-driven models resembles the base-pairing structure formed by nitrogenous bases in \textit{DNA}; and the hierarchical structure also coincides with the spiraling progression of the \textit{DNA}. Under the proposed architecture, we further introduce two distinct interaction strategies between mechanistic and data-driven models, enabling effective bidirectional interaction between the two components. In addition, we also propose two corresponding data-driven frameworks, which, together with the hierarchical structure of DNA, facilitate multi-stage modeling for complex industrial systems. The main contributions of our work are as follows:
\begin{itemize}[leftmargin=2.5em]
    \item We propose a novel hybrid modeling architecture, DNA, in which data-driven models permanently embed their enhancements into the mechanistic models, enabling the resulting models to be decoupled from the data-driven model and independently deployed in industrial practice.


    \item We design two interaction strategies for different types of mechanistic models and propose two novel data-driven frameworks that are aligned with the above interaction strategies.
    \item Extensive experiments on two distinct industrial systems demonstrate the effectiveness and robustness of our DNA architecture. Another merit of DNA is that improvements in upstream stages can propagate downstream, leading to more accurate subsequent predictions.
\end{itemize}


\section{Task Definition and Architecture Overview}

\subsection{Task Definition}
\label{task definition}
A key step in building an industrial digital twin is to construct a high-fidelity model $\mathcal{M}$ of the target industrial system $\mathcal{O}$. Given the same input conditions $X$ as the real-world system, the model should produce outputs that closely match the real outputs. In this way, the model can simulate the behavior of the physical system and serve as its digital twin. This task can be formulated as: $\min_\mathcal{M} \; \mathbb{E}_{X \sim \mathcal{X}} \left[ \left| \mathcal{M}(X) - \mathcal{O}(X) \right| \right]$, where $\mathbb{E}[\cdot]$ denotes the mathematical expectation and $\mathcal{X}$ denotes the distribution of operating conditions. Three-phase pipelines are mainly used to transport mixtures of oil, gas, and water and play an important role in offshore oil and gas production \cite{nwanwe2022comparison}. Here, we use it as a representative example to illustrate the proposed approach.





We first construct an original mechanistic model $\mathcal{M}_{me}$ for the three-phase pipeline based on the Beggs-Brill theory \cite{beggs1973study,chaves2022comparative}. The model contains two sequential modeling stages. In Stage I, the mechanistic model $\mathcal{M}_{me}^{I}$ uses the original operating conditions ${X}$ to calculate the physical properties of the different phases, denoted as ${Y}_{{me}}^{I}$. In Stage II, $\mathcal{M}_{{me}}^{II}$ takes $X,{Y}_{{me}}^{I}$ as input and performs further calculations to obtain the outputs ${Y}_{{me}}^{II}$ for the two downstream variables. The detailed data flow is provided in Appendix \ref{dataflow}. Following the Beggs-Brill theory, $\mathcal{M}_{{me}}^{I}$ mainly consists of explicit functions: $y = f(x_1, x_2, \ldots, x_n)$ and can be written as $\mathcal{M}_{{me}}^{I}=\{f_1,f_2,\ldots,f_n\}$. In contrast, $\mathcal{M}_{{me}}^{II}$ mainly consists of implicit equations: $F(x_1,x_2,\ldots,x_n,y,\dot{y},\ldots)=0$ and can be written as $\mathcal{M}_{{me}}^{II}=\{F_1,F_2,\dots,F_n\}$, where $\dot{y}=\frac{dy}{dt}$ denotes the time derivative of $y$.

\subsection{Architecture Overview}

\begin{figure}
\centering
\includegraphics[width=\textwidth]{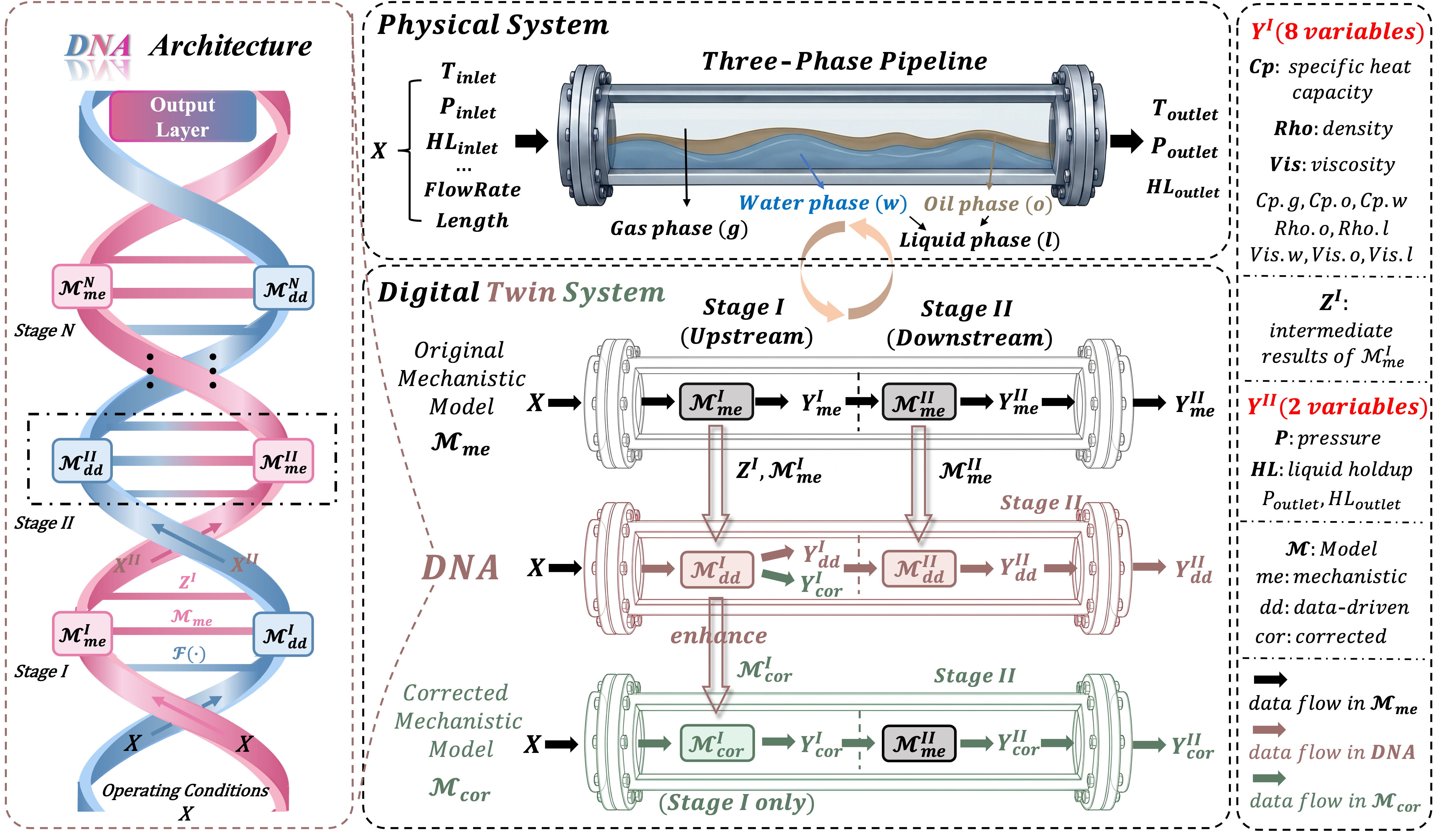}
\caption{Task demonstration and DNA architecture. Left: DNA architecture; middle: the three-phase pipeline digital twin system; right: definitions of the variables. }
\label{fig1}
\end{figure}

The hierarchical structure of DNA is designed for complex industrial systems that require multi-stage modeling. As shown in Fig.\ref{fig1}, $\mathcal{M}_{{me}}^{I}$ and $\mathcal{M}_{{me}}^{II}$ are sequentially connected along the mechanistic-model chain (red), while the corresponding data-driven models, $\mathcal{M}_{{dd}}^{I}$ and $\mathcal{M}_{{dd}}^{II}$, form the data-driven-model chain (blue). Together, they form the DNA architecture. DNA provides two interaction strategies between mechanistic and data-driven models.

In Stage I, $\mathcal{M}_{{dd}}^{I}$ takes the operating conditions ${X}$, intermediate results ${Z^I}$ of $\mathcal{M}_{{me}}^{I}$, and mechanistic model $\mathcal{M}_{{me}}^{I}$ as inputs, where $\mathcal{M}_{{me}}^{I}$ provides physical prior knowledge. $\mathcal{M}_{{dd}}^{I}$ simultaneously generates predictions ${Y}_{{dd}}^{I}$ and coefficients ${coefs}$, which are used to construct a correction function $\mathcal{F}(\cdot)$. The correction function is fed back to $\mathcal{M}_{{me}}^{I}$ to refine its predictions, yielding the enhanced model $\mathcal{M}_{{cor}}^{I}$ and its outputs ${Y}_{cor}^{I}$. We call this interaction a bidirectional asynchronous compensation strategy. In Stage II, $\mathcal{M}_{{dd}}^{II}$ incorporates prior knowledge of $\mathcal{M}_{{me}}^{II}$ through graph neural networks (GNNs), allowing the mechanistic and data-driven components to function as a unified model and jointly produce the predictions $Y_{dd}^{II}$. We call this interaction a mechanism-embedding strategy. The green pipeline in Fig.\ref{fig1} is a control configuration for the original mechanistic model (black). Only $\mathcal{M}_{{me}}^{I}$ is replaced by the enhanced model $\mathcal{M}_{{cor}}^{I}$, while the remaining components are unchanged. This configuration evaluates the gain propagation capability of DNA: whether the improvement obtained at the upstream stages can propagate to downstream stages such that subsequent modeling stages also benefit from the upstream enhancement.

In particular, DNA is an industrial modeling architecture rather than merely a specific model. Within the DNA architecture, the data-driven model $\mathcal{M}_{{dd}}^{i}$ at each stage $i\in\{1,2,\dots,N\}$ can be replaced by any suitable existing deep learning method. Thus, DNA provides a general architecture for integrating mechanistic and data-driven models across multiple modeling stages.

\section{Methodology}

\begin{figure}
\centering
\includegraphics[width=\textwidth]{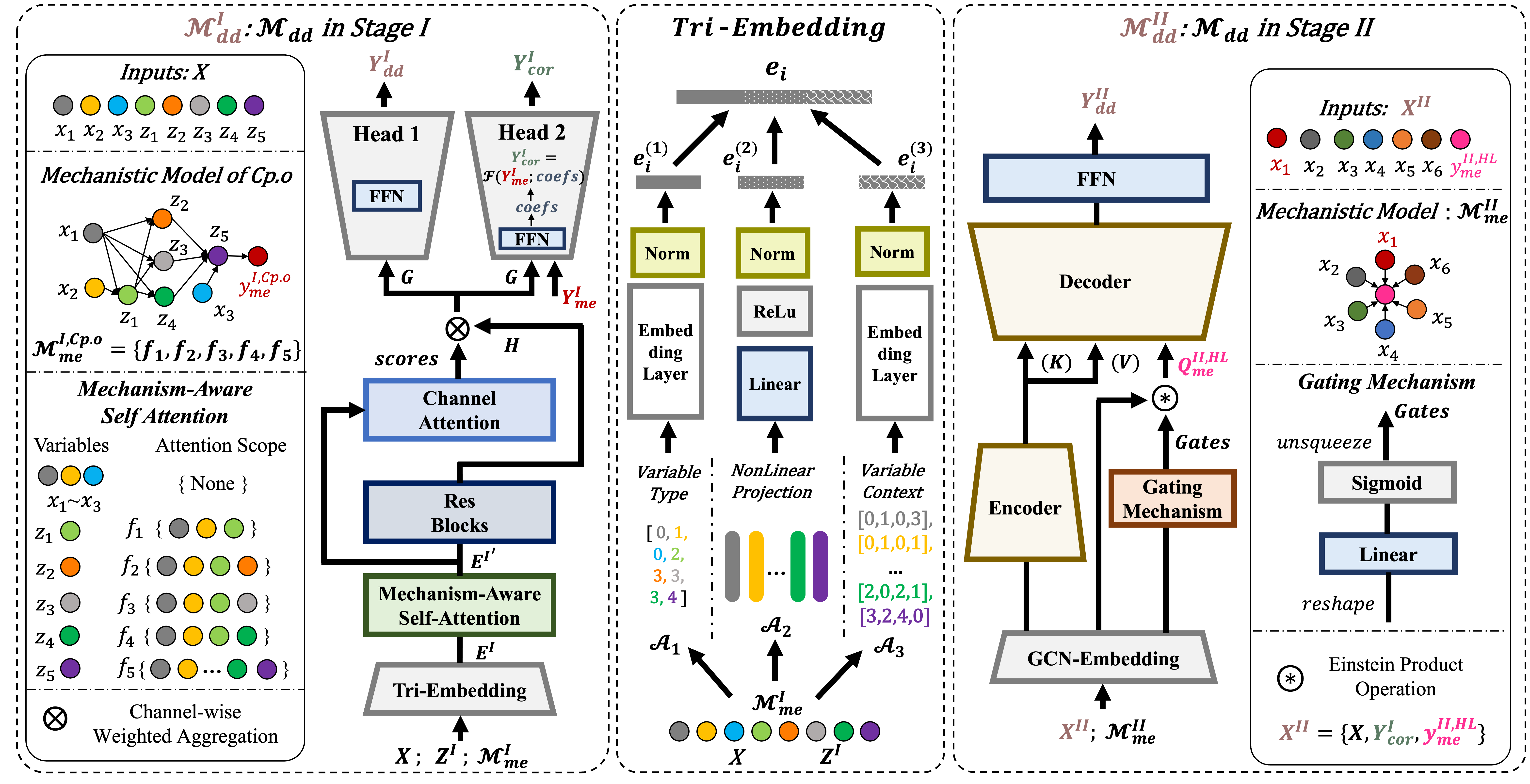}
\caption{Framework of the proposed data-driven modeling methods and Tripartite Embedding (Tri-Embedding). Left: a concrete \(Cp.o\) case is used to illustrate the topology of the mechanistic model in Stage I, the logic of the proposed Mechanism-Aware Self-Attention (MASA), and the overall framework of \(\mathcal{M}_{dd}^{I}\). Middle: the workflow of the proposed Tri-Embedding, illustrated using the same \(Cp.o\) case. Right: $HL$ is used as an case to illustrate the topology of the mechanistic model in Stage II, the proposed gating mechanism, and the overall framework of $\mathcal{M}_{dd}^{II}$.}
\label{fig2}
\end{figure}

In addition to the DNA architecture and two interaction strategies, this work also develops two data-driven frameworks tailored to these strategies, as shown in Fig.\ref{fig2}.

\subsection{Data-Driven Framework in Stage \Rmnum{1}}

\paragraph{Tri-Embedding.} Industrial datasets often comprise both observable variables (e.g., temperature, pressure, flow rate) and latent ones (e.g., specific heat capacity, density, viscosity) \cite{koretsky2012engineering}, all represented as scalar real values. Our datasets follow the same convention. Relying solely on such low-dimensional inputs to model complex industrial systems is clearly insufficient. Therefore, effective methods for dimensional enhancement are crucial. When the mechanistic model of the industrial system consists of a large number of functions with strongly nested and coupled dependencies, such as $\mathcal{M}^{I,Cp.o}_{me}$ in Fig.\ref{fig2}, it can result in extremely complex topological graphs. Although graph-based methods, for example graph convolutional networks (GCNs) \cite{kipf2016semi}, are effective in generating high-dimensional representations, their direct application to highly complex topologies can substantially increase the computational burden. To address this issue, we propose the tripartite embedding (Tri-Embedding), which involves only embedding lookups and a simple nonlinear projection, resulting in minimal computational overhead. The Tri-Embedding process consists of three components: \textbf{Variable Type Encoding ($\mathcal{A}_1$)}; \textbf{Nonlinear Projection ($\mathcal{A}_2$)}; \textbf{Variable Context Encoding ($\mathcal{A}_3$)}. Each component will generate their own representation $(e_i^{(1)},e_i^{(2)},e_i^{(3)})$, then we concatenate all three parts to obtain the full high dimensional representation of variable $i$, $e_i=CAT(e_i^{(1)},e_i^{(2)},e_i^{(3)})$. Further implementation detail of Tri-Embedding can be found in Appendix \ref{Tri-Embedding}. In the mechanistic model, the output variable $Y^I_{me}$ is the final computed result and does not participate in the internal computation process. To maintain consistency with the mechanistic model, we exclude $Y^I_{me}$ from the input $X^I$ of $\mathcal{M}^I_{dd}$ as well. The process of generating a high-dimensional representation $E^I$ for each variable is formulated as follows:$E^I=T(X^I;\mathcal{M}^I_{me}), E^I \in \mathbb{R}^{n \times d_m},$
\noindent where $T(\cdot)$ denotes the Tri-Embedding operation, $X^I =\{X,Z^I\} =\{x_1, \ldots, x_{n_1}, z_1, \ldots, z_{n_2}\} \in \mathbb{R}^{n \times 1}$, $E^I = \{e_1, e_2, \ldots, e_n\} \in \mathbb{R}^{n \times d_m}$, $n = n_1 + n_2$ is the total number of variables in the input, and $d_m=32$ is the embedding dimension, as specified in the Dim column of Tab.\ref{table5}.

\paragraph{Data-Driven Framework.} The self-attention mechanism is well suited to our setting, as it can capture dependencies among variables regardless of their relative topological distance, which aligns with the complex topology of mechanistic models where relevant variables may be far apart. However, as illustrated by $\mathcal{M}^{I,Cp.o}_{me}$ in Fig.\ref{fig2}, the computation of intermediate physical quantities $\{z_i\}_{i=1}^5$ is inherently directional. For example, $z_1$ depends only on $x_1$ and $x_2$, and is independent of all other variables. In other words, according to the underlying mechanism of the industrial system, variables beyond $x_1$ and $x_2$ should not influence $z_1$. Therefore, vanilla self-attention, which computes attention weights across all input tokens indiscriminately, contradicts the real-world mechanistic constraints in our digital twin setting. To address this issue, we propose Mechanism-Aware Self-Attention (MASA). MASA assigns a dedicated attention scope to each $z_i$ based on $\mathcal{M}^I_{me}$, restricting its attention to variables that are mechanistically relevant to itself. This design offers two key benefits: (i) it enforces mechanistic consistency in $\mathcal{M}^I_{dd}$; (ii) it eliminates unnecessary attention operations, reducing computational cost, as shown in Fig.\ref{fig3}d. Given the high-dimensional input representation $E^I$, MASA produces updated representation $E^{I^\prime}$ that aggregate more informative features, as formulated below:
\begin{equation}
    E^{I^\prime}={MASA}(E^I,E^I,E^I),E^{I^\prime}\in \mathbb{R}^{n \times d_m}.
    \label{equa-4}
\end{equation}
Moreover, in mechanistic model, different input variables may contribute unequally to the final output $Y^I_{me}$. Treating all variables identically during prediction can therefore be suboptimal. Inspired by the channel attention mechanism \cite{woo2018cbam}, we introduce a channel attention module that assigns each variable an importance score, reflecting its contribution to the prediction. Specifically, we first add an extra dimension to $E^{I^\prime}$ to form $E^{I^{\prime}}_{exp}$, which is then fed into three ResBlocks \cite{he2016deep} to extract deeper features $H$. In parallel, $E^{I^\prime}_{exp}$ is processed by the channel attention module to produce a $scores$ vector whose length matches the number of input variables. $scores$ are then used to compute a weighted aggregation of $H$ along the channel dimension, resulting in the fused representation $G$. More details about ResBlocks and channel attention module can be found in Appendix \ref{res} and \ref{Channel}. These processes are formulated as:
\begin{equation}
    E^{I^\prime}_{exp} = unsqueeze(E^{I^\prime},dim=-1),E^{I^\prime}_{exp}\in\mathbb{R}^{n\times d_m\times 1},H=\mathcal{R}(E^{I^\prime}_{exp}),H\in\mathbb{R}^{n\times d_m\times 1},
    \label{equa-2}
\end{equation}
\begin{equation}
scores=\sigma(L(GAP(E^{I^\prime}_{exp}))),scores\in\mathbb{R}^{n\times 1\times 1},G = H\otimes scores,G\in\mathbb{R}^{1\times d_m\times 1},
    \label{equa-7}
\end{equation}
\noindent where $unsqueeze(\cdot)$ denotes the dimension expansion operation; $\mathcal{R}(\cdot)$ denotes ResBlocks; $GAP(\cdot)$ represent global average pooling; $L(\cdot)$ denotes a series of linear transformations with $ReLU$ activations \cite{nair2010rectified}; $\sigma(\cdot)$ is the $Sigmoid$ function that generates channel-wise weights; and $\otimes$ is the channel-wise weighted aggregation. Finally, the fused feature $G$ is fed into two prediction heads to generate the data-driven prediction $Y^I_{{dd}}$ and the correction function $\mathcal{F}(\cdot)$. This process is formulated as follows:
\begin{equation}
    Y^I_{dd}=Head_1(squeeze(G)),coefs=Head_2(squeeze(G)),Y^I_{cor}=\mathcal{F}(Y^I_{me};coefs),
    \label{equa-9}
\end{equation}
\noindent here, $squeeze(\cdot)$ is used to remove redundant dimensions; $Head_1(\cdot)$ is a two-layer feedforward network, while $Head_2(\cdot)$ is a three-layer feedforward network that produces a set of coefficients $coefs$ to construct $\mathcal{F}(\cdot)$ to refine the original mechanistic model's outputs $Y^I_{me}$ .

\subsection{Data-Driven Framework in Stage \Rmnum{2}}
\paragraph{GCN-Embedding.} When the mechanistic model of an industrial system is formulated as algebraic or differential equations rather than explicit functions, the modeling task becomes one of predicting the unknown variables within these equations. Stage II is designed to handle such cases. In this case, since variables other than the unknowns are typically either computed in earlier stages or inherently known, the topological graphs of the mechanistic model $\mathcal{M}^{II}_{me}$ in this stage are typically simpler than those of $\mathcal{M}^I_{me}$ in Stage I, as shown in Fig.\ref{fig2}. This relatively simple topology alleviates concerns about the computational overhead associated with GCNs. Moreover, we aim to provide a more diverse data-driven strategy for digital twin modeling. These considerations motivate us to adopt GCNs as an alternative approach for generating high-dimensional representations of the input $X^{II}$. In addition, a key difference here is that since the target variable $y^{II}_{me}$ itself participates in the equation-solving process,  we additionally include $y^{II}_{me}$ as part of the input to $\mathcal{M}^{II}_{dd}$.
\begin{equation}
    E^{II}=GCN(X^{II};\mathcal{G}), E^{II}\in\mathbb{R}^{(n+1)\times d_m},
    \label{equa-13}
\end{equation}
\noindent where $X^{II}=\{x_1,\ldots,x_n,y^{II}_{{me}}\}\in\mathbb{R}^{(n+1)\times 1}$, $E^{II}=\{e_1,\ldots,e_n,e_{y^{II}_{{me}}}\}\in\mathbb{R}^{(n+1)\times d_m}$, and $\mathcal{G}$ denotes the topological graph of $\mathcal{M}^{II}_{me}$.

\paragraph{Data-Driven Framework.} Encoder–Decoder architectures have demonstrated remarkable performance across a wide range of tasks due to their strong ability to effectively capture complex relationships between inputs and outputs. We adopt the Transformer backbone in our design, with the key distinction lying in the decoder. Detailed architectures of the encoder $Enc(\cdot)$ and decoder $Dec(\cdot)$ are provided in the Appendix \ref{enc-dec}. The encoder component largely follows the standard Transformer encoder architecture, and the encoding process is formulated as $O_{en}=Enc(E^{II},E^{II},E^{II})$, where $O_{en}\in \mathbb{R}^{(n+1)\times d_m}$. The decoder component only includes the cross-attention module of a standard Transformer and all subsequent layers. Since $y^{II}_{me}$ originates from the mechanistic model prediction, it inherently carries physical or domain-specific information and remains highly correlated with the ground truth $y^{II}_{true}$. Therefore, we use $y^{II}_{{me}}$ as the query to guide the attention mechanism. However, this design relies on the quality of $y^{II}_{me}$. When the quality of mechanistic model’s output is poor, $y^{II}_{me}$ may guide attention toward misleading patterns, thus exacerbating the propagation of errors. To mitigate this risk, we further introduce a gating mechanism $GM(\cdot)$ that generates $Gates$ for input variables and uses the Einstein product $\circledast$ to compute their weighted sum as the final query $Q_{me}$, as illustrated in Stage \Rmnum{2} of Fig.\ref{fig2}. This process can be formulated as follows:
\begin{equation}
    Gates=GM(E^{II}),Gates \in \mathbb{R}^{(n+1)\times 1},Q_{me}=E^{II}\circledast Gates,Q_{me} \in \mathbb{R}^{1 \times d_m}.
    \label{equa-14}
\end{equation}
\noindent Finally, decoder performs cross-attention using $Q_{{me}}$, $O_{{en}}$, and $O_{{en}}$ as the query($Q$), key($K$), and value($V$), respectively. Its output is then passed through a three-layer feedforward network $FFN(\cdot)$ to predict the final output $Y^{II}_{{dd}}$, as formulated below:
\begin{equation}
    Y^{II}_{dd}=FFN(Dec(Q_{me},O_{en},O_{en})).
    \label{equa-17}
\end{equation}

Details of the model optimization procedure are provided in the Appendix \ref{Model Optimization}.

\section{Experiment}

\subsection{Industrial Systems}
\label{datasets}

\textbf{Three-phase pipeline.} As described in Section \ref{task definition}, we use the three-phase pipeline as the primary industrial system in our experiments. We first collect real-world data under 759 distinct operating conditions across 20 locations, including the operating conditions and measurements of ten variables, comprising eight physical properties in Stage I and two downstream variables in Stage II, as shown in Fig.\ref{fig1}. The operating conditions serve as the inputs ($X$), while the measurements of the ten variables serve as the ground-truth outputs ($Y^I_{true},Y^{II}_{true}$). We then simulate the mechanistic model under the same conditions to obtain mechanistic predictions ($Y^I_{me},Y^{II}_{me}$) and intermediate variables $Z^I$. These data yield eight datasets for Stage \Rmnum{1} and two for Stage \Rmnum{2}. In total, the mechanistic model generates 15,180 samples. More details about three-phase pipeline are provided in Appendix \ref{Modeled Industrial System}.

\textbf{Acentric factor.} Acentric factor is a key parameter that characterizes the extent to which real gases deviate from ideal gas behavior, playing an important role in thermodynamics and fluid mechanics \cite{biswas2023predicting}. Unlike tangible industrial equipment(three-phase pipeline), the acentric factor represents an abstract variable that is difficult to measure directly yet remains critical to industrial processes. Therefore, the three-phase pipeline and the acentric factor correspond to fundamentally different types of industrial object in real-world scenarios. Accordingly, we select the acentric factor as the second experimental target to further evaluate the generalization capability of DNA. More details about acentric factor are provided in the Appendix \ref{Modeled Industrial System: Acentric Facotr}.

\subsection{Baselines}

Our tasks can generally be categorized as regression problems. Accordingly, we consider four groups of baselines that have shown strong performance in regression tasks, including (\romannumeral1) Tree-based  method: LightGBM ~\cite{ke2017lightgbm}; (\romannumeral2) Transformer-based methods: Transformer ~\cite{vaswani2017attention}, FlowformerMH ~\cite{wu2022flowformerlinearizingtransformersconservation}, iTransformer ~\cite{liu2024itransformerinvertedtransformerseffective}, and MultiPatchFormer (MPFormer)\cite{naghashi2025multiscale}; (\romannumeral3) Graph-based methods: GATv2 ~\cite{velivckovic2017graph}, DGDL ~\cite{zhu2022dynamic}, TGCN-S ~\cite{kong2023collaborative}, and RaDA ~\cite{chen2024residual}; (\romannumeral4) Other types of models: TimesNet ~\cite{wu2022timesnet}, TimeXer \cite{wang2024timexer} and UMixer ~\cite{ma2024umixerunetmixerarchitecturestationarity}. Detailed descriptions of the baselines and the rationale for their selection are provided in the Appendix \ref{baselines}.

\subsection{Overall Performance}

\begin{table*}[]
\centering
\caption{Overall performance on the two modeling task. \textbf{\textcolor{red}{Red}}: the best, \textcolor{blue}{\underline{Blue}}: the 2nd best.}

\Huge
\resizebox{\textwidth}{!}{%
{
\begin{tabular}{c|cc|cc|cc|cc|cc|cc|cc|cc|cc|cc|cc|cc|cc|cc}

\specialrule{3pt}{0pt}{3pt}

\multicolumn{3}{c|}{Category}
& \multicolumn{2}{c|}{\includegraphics[height=1.2em]{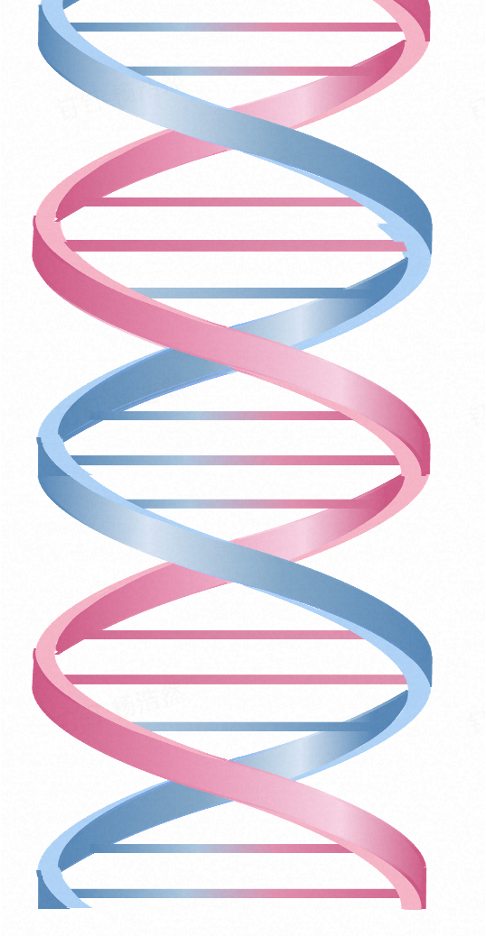}\,\textbf{\Huge{DNA}}}
& \multicolumn{2}{c|}{\includegraphics[height=1.2em]{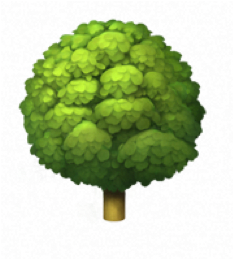}\,\textbf{\Huge{Tree-based}}}
& \multicolumn{8}{c|}{\includegraphics[height=1.2em]{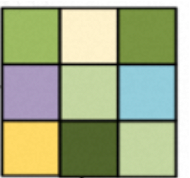}\,\textbf{\Huge{Transformer-based}}}
& \multicolumn{8}{c|}{\includegraphics[height=1.2em]{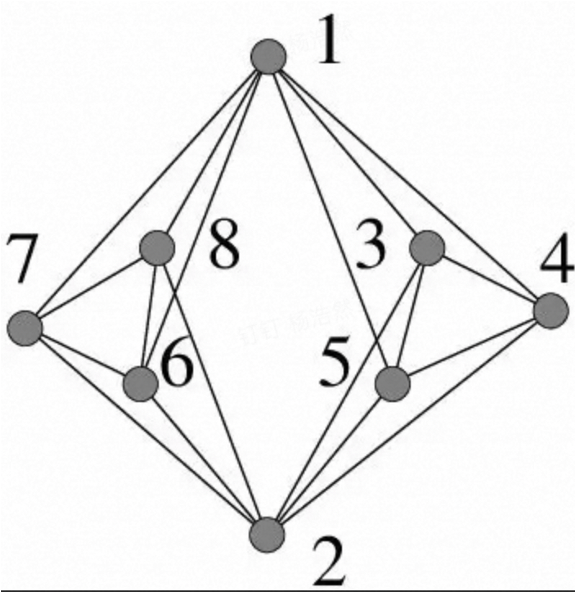}\,\Huge{\textbf{Graph-based}}}
& \multicolumn{6}{c}{\includegraphics[height=1.2em]{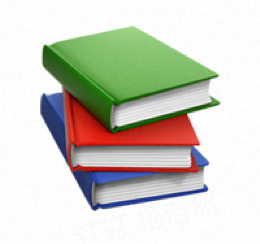}\,\Huge{\textbf{Others}}}
\\
\midrule

\multicolumn{3}{c|}{Model}
& \multicolumn{2}{c|}{\textbf{Ours}}
& \multicolumn{2}{c|}{LightGBM}
& \multicolumn{2}{c|}{Transformer}
& \multicolumn{2}{c|}{FlowformerMH}
& \multicolumn{2}{c|}{iTransformer}
& \multicolumn{2}{c|}{MPFormer}
& \multicolumn{2}{c|}{GATv2}
& \multicolumn{2}{c|}{DGDL}
& \multicolumn{2}{c|}{TGCNS}
& \multicolumn{2}{c|}{RaDA}
& \multicolumn{2}{c|}{TimesNet}
& \multicolumn{2}{c|}{TimeXer}
& \multicolumn{2}{c}{Umixer}
\\
\midrule

\multicolumn{3}{c|}{Metric}                           & MAE           & \multicolumn{1}{c|}{RMSE}         & MAE           & \multicolumn{1}{c|}{RMSE}          & MAE             & \multicolumn{1}{c|}{RMSE}           & MAE         & \multicolumn{1}{c|}{RMSE}        & MAE             & \multicolumn{1}{c|}{RMSE}            & MAE          & \multicolumn{1}{c|}{RMSE}        & MAE          & \multicolumn{1}{c|}{RMSE}        & MAE             & \multicolumn{1}{c|}{RMSE}            & MAE          & \multicolumn{1}{c|}{RMSE}         & MAE   & \multicolumn{1}{c|}{RMSE} & MAE   & \multicolumn{1}{c|}{RMSE} & MAE   & \multicolumn{1}{c|}{RMSE} & MAE   & \multicolumn{1}{c}{RMSE} \\ \specialrule{3pt}{3pt}{3pt}

\multicolumn{1}{c|}{\multirow{36}{*}{\rotatebox{90}{\textbf{Stage \Rmnum{1}}}}}  & \multicolumn{1}{c|}{\multirow{4}{*}{\rotatebox{90}{Cp.g}}} & MM  & 49.58       & 50.72    & 49.58       & 50.72    & 49.58       & 50.72    & 49.58       & 50.72    & 49.58       & 50.72    & 49.58       & 50.72    & 49.58       & 50.72    & 49.58       & 50.72    & 49.58       & 50.72    & 49.58       & 50.72    & 49.58       & 50.72    & 49.58       & 50.72    & 49.58       & 50.72      \\

& \multicolumn{1}{c|}{}  & DD  & \textbf{\textcolor{red}{0.763}}       & 1.073               & 0.913         & 1.205         & 1.754           & 2.013          & 1.467       & 1.965       & 1.037           & 1.312           & 0.795        & 1.073      & 0.851        & 1.094       & 1.387           & 1.736           & 0.832        & \textcolor{blue}{\underline{1.065}}        & 1.300 & 1.619      &\textcolor{blue}{\underline{0.794}} & \textbf{\textcolor{red}{1.060}}  & 0.807  &1.068 & 0.843 &1.100               \\

& \multicolumn{1}{c|}{}                       & Cor & 0.831       & \textcolor{blue}{\underline{1.067}}      & \rule[0.07cm]{0.8cm}{1pt}              & \rule[0.07cm]{0.8cm}{1pt}          & 1.253           & 1.943          &0.780       & 1.111       & 1.130           & 1.747           & 0.812        & 1.120       & \textcolor{blue}{\underline{0.770}}        & \textbf{\textcolor{red}{1.051}}       & 0.799           & 1.143           & 0.823       & 1.101        & \textbf{\textcolor{red}{0.767}} & 1.073      & 0.792 & 1.138  &0.859 &1.130 & 0.792 & 1.069              \\

\rowcolor{gray!20}
& \multicolumn{1}{c|}{}                       & Avg     & \textcolor{blue}{\underline{0.797}}   & \textbf{\textcolor{red}{1.070}}           & \rule[0.07cm]{0.8cm}{1pt}         & \rule[0.07cm]{0.8cm}{1pt}         & 1.503           & 1.978          & 1.123       & 1.538       & 1.083           & 1.530           & {0.804}        & 1.096       & 0.811        & \textcolor{blue}{\underline{1.073}}       & 1.093           & 1.440           & 0.827        & 1.083        & 1.034 & 1.346    & \textbf{\textcolor{red}{0.793}}   &  1.099 & 0.833 & 1.099& 0.817 & 1.084                \\  \cmidrule(lr){2-29}

& \multicolumn{1}{c|}{\multirow{4}{*}{\rotatebox{90}{Cp.o}}}  & MM  & 15.01     & 16.03      & 15.01     & 16.03    & 15.01     & 16.03    & 15.01     & 16.03    & 15.01     & 16.03    & 15.01     & 16.03    & 15.01     & 16.03    & 15.01     & 16.03    & 15.01     & 16.03    & 15.01     & 16.03       & 15.01     & 16.03  & 15.01     & 16.03 & 15.01     & 16.03   \\

&\multicolumn{1}{c|}{}   & DD  & \textcolor{blue}{\underline{1.704}}   & \textbf{\textcolor{red}{2.335}}                & 2.114         & 2.789         & 1.716           & 2.402          & 2.003       & 2.578       & 1.803           & 2.438           & \textbf{\textcolor{red}{1.673}}        & 2.416       & 1.730        & 2.394       & 2.139           & 2.695           & 2.225       & 2.742        & 2.419 & 3.044      &1.728 & \textcolor{blue}{\underline{2.368}} & 1.837 &2.476 &1.733    &2.420               \\

&\multicolumn{1}{c|}{}                       & Cor & 1.895       & 2.474       & \rule[0.07cm]{0.8cm}{1pt}          & \rule[0.07cm]{0.8cm}{1pt}                & 1.958           & 2.517          & 1.893       & 2.509       & 1.871           & 2.515           & 1.936        & 2.491       & 1.894        & 2.511       & \textcolor{blue}{\underline{1.892}}           & 2.488         & 1.910        & 2.500        & 1.893 &  \textcolor{blue}{\underline{2.473}}    & 1.906 & 2.510 &    \textbf{\textcolor{red}{1.880}}     &2.516      &1.895      &\textbf{\textcolor{red}{2.457}}            \\

\rowcolor{gray!20}
&\multicolumn{1}{c|}{}                       & Avg & \textbf{\textcolor{red}{1.799}}       & \textbf{\textcolor{red}{2.404}}       & \rule[0.07cm]{0.8cm}{1pt}         & \rule[0.07cm]{0.8cm}{1pt}                  & 1.837           & 2.460          & 1.948       & 2.543       & 1.837           & 2.477           & \textcolor{blue}{\underline{1.805}}        & 2.454       & 1.812        & 2.453       & 2.016           & 2.592           & 2.068        & 2.621        & 2.156   & 2.759       &  1.817   &    2.439     &1.859  &2.496  &1.814      &\textcolor{blue}{\underline{2.438}}           \\ \cmidrule(lr){2-29}

&\multicolumn{1}{c|}{\multirow{4}{*}{\rotatebox{90}{Cp.w}}}  &MM &  2.354  & 2.898  &  2.354  & 2.898 &  2.354  & 2.898 &  2.354  & 2.898 &  2.354  & 2.898 &  2.354  & 2.898 &  2.354  & 2.898 &  2.354  & 2.898 &  2.354  & 2.898 &  2.354  & 2.898 &  2.354  & 2.898 &  2.354  & 2.898 &  2.354  & 2.898 \\

&\multicolumn{1}{c|}{}   & DD  & 0.724       & 1.017     &1.929  &2.390    & 2.692         & 2.864        & 1.799         & 1.985         & 1.084           & 1.437          & 0.677       & 1.027       & 1.674           & 1.851           & 1.151        & 1.519       & 0.874        & 1.128       & 0.850           & 1.225           & \textbf{\textcolor{red}{0.553}}        & \textbf{\textcolor{red}{0.835}}        &  \textcolor{blue}{\underline{0.668}}  & \textcolor{blue}{\underline{0.936}}    &0.774 & 1.054                 \\

&\multicolumn{1}{c|}{}                       & Cor & \textbf{\textcolor{red}{1.219}}       &\textbf{\textcolor{red}{1.611}}      & \rule[0.07cm]{0.8cm}{1pt}          & \rule[0.07cm]{0.8cm}{1pt}                  & 1.761           & 2.183          & 1.675       & 2.141       & 1.685           & 2.143           & 1.723        & 2.201       & \textcolor{blue}{\underline{1.630}}        & \textcolor{blue}{\underline{2.068}}       & 1.754           & 2.185           & 1.716        & 2.222        & 1.732 & 2.198     & 1.701  & 2.150 &1.664      &2.098      &1.685      &2.129                \\

\rowcolor{gray!20}
&\multicolumn{1}{c|}{}                       & Avg & \textbf{\textcolor{red}{0.972}}       & \textbf{\textcolor{red}{1.314}}      & \rule[0.07cm]{0.8cm}{1pt}         & \rule[0.07cm]{0.8cm}{1pt}            & 2.226           & 2.523          & 1.737       & 2.063       & 1.384           & 1.790           & 1.200        & 1.614       & 1.652        & 1.960       & 1.452           & 1.852           & 1.295        & 1.675        & 1.291     & 1.711    & \textcolor{blue}{\underline{1.127}}        &\textcolor{blue}{\underline{1.492}}       &1.166      &1.517      &1.229      &1.592           \\ \cmidrule(lr){2-29}

&\multicolumn{1}{c|}{\multirow{4}{*}{\rotatebox{90}{Rho.o}}} &MM    & 13.93   & 14.15     & 13.93   & 14.15     & 13.93   & 14.15     & 13.93   & 14.15     & 13.93   & 14.15     & 13.93   & 14.15     & 13.93   & 14.15     & 13.93   & 14.15     & 13.93   & 14.15     & 13.93   & 14.15     & 13.93   & 14.15     & 13.93   & 14.15    & 13.93   & 14.15   \\

&\multicolumn{1}{c|}{}   & DD  &\textbf{\textcolor{red}{0.793}}      & \textbf{\textcolor{red}{1.367}}            & 1.245         & 1.915         & 0.956           & 1.473          & 1.537       & 1.852       & \textcolor{blue}{\underline{0.824}}           & 1.401           & 1.021        & 1.676       & 1.469        & 1.793       & 1.332           & 1.765           & 1.644        & 2.038        & 0.838         & 1.403       & 1.045       &    1.707     &1.097      &1.636      &0.845      &\textcolor{blue}{\underline{1.392}}           \\

&\multicolumn{1}{c|}{}                       & Cor & \textcolor{blue}{\underline{1.504}}       & 2.397      & \rule[0.07cm]{0.8cm}{1pt}                  & \rule[0.07cm]{0.8cm}{1pt}          & 1.532           & 2.385          & 1.557       & 2.394       & 1.595           & \textbf{\textcolor{red}{2.381}}           & 1.571        & \textcolor{blue}{\underline{2.382}}       & 1.625        & 2.417       & 1.576           & 2.390           & 1.591        & 2.390        & 1.552        & 2.388    & 1.779          & 2.430        & \textbf{\textcolor{red}{1.486}}   &2.398      &1.601      &2.384                \\

\rowcolor{gray!20}
&\multicolumn{1}{c|}{}                       & Avg & \textbf{\textcolor{red}{1.148}}      & \textbf{\textcolor{red}{1.882}}    & \rule[0.07cm]{0.8cm}{1pt}        & \rule[0.07cm]{0.8cm}{1pt}             & 1.244           & 1.929          & 1.547       & 2.123       & 1.210           & 1.891           & 1.296        & 2.029       & 1.547        & 2.105       & 1.454           & 2.077           & 1.617        & 2.214        &\textcolor{blue}{\underline{1.195}}  & 1.896     &  1.412   &  2.068   &1.291      &2.017   &1.223    & \textcolor{blue}{\underline{1.888}}               \\ \cmidrule(lr){2-29}

&\multicolumn{1}{c|}{\multirow{4}{*}{\rotatebox{90}{Rho.l}}} & MM    & 23.07  & 26.64  & 23.07  & 26.64  & 23.07  & 26.64  & 23.07  & 26.64  & 23.07  & 26.64  & 23.07  & 26.64  & 23.07  & 26.64  & 23.07  & 26.64  & 23.07  & 26.64  & 23.07  & 26.64  & 23.07  & 26.64   & 23.07  & 26.64  & 23.07  & 26.64\\

&\multicolumn{1}{c|}{}    & DD  & \textcolor{blue}{\underline{2.434}}       & \textbf{\textcolor{red}{3.437}}              & 3.204         & 4.019         & 2.696           & 3.576          & 3.998       & 4.556       & 2.523         & 3.681          & 3.178        & 4.249       & 4.345        & 4.859       & 3.078           & 4.118           & 3.202        & 4.098        & 3.483       & 4.190        & 2.748 &    3.846      &3.137      &4.534      &\textbf{\textcolor{red}{2.259}}     & \textcolor{blue}{\underline{3.519}}     \\

&\multicolumn{1}{c|}{}                       & Cor & \textcolor{blue}{\underline{6.583}}     &7.946      & \rule[0.07cm]{0.8cm}{1pt}                 & \rule[0.07cm]{0.8cm}{1pt}          & 6.643           & 7.990          & \textbf{\textcolor{red}{6.564}}      &  \textcolor{blue}{\underline{7.941}}      & 6.754           & 8.153           & 6.722        & 8.120       & \textbf{\textcolor{red}{6.564}}        & \textbf{\textcolor{red}{7.938}}       & 6.683           & 8.036           & 6.646        & 8.000        & 6.590        & 7.953      &  6.957        & 8.445           & 6.771     &8.203      &6.776      &8.135     \\

\rowcolor{gray!20}
&\multicolumn{1}{c|}{}                       & Avg & \textbf{\textcolor{red}{4.509}}      & \textbf{\textcolor{red}{5.692}}     & \rule[0.07cm]{0.8cm}{1pt}       & \rule[0.07cm]{0.8cm}{1pt}               & 4.670           &  \textcolor{blue}{\underline{5.783}}         & 5.281      & 6.248       & 4.638           & 5.917           & 4.950        & 6.185       & 5.454        & 6.399       & 4.880           & 6.077           & 4.924        & 6.049        & 5.036      & 6.072        & 4.852    &   6.146           &4.954       &6.368  &\textcolor{blue}{\underline{4.517}}        & 5.827       \\ \cmidrule(lr){2-29}

&\multicolumn{1}{c|}{\multirow{4}{*}{\rotatebox{90}{Vis.w}}} & MM   & 9.239 & 15.45  & 9.239 & 15.45  & 9.239 & 15.45    & 9.239 & 15.45    & 9.239 & 15.45    & 9.239 & 15.45    & 9.239 & 15.45    & 9.239 & 15.45    & 9.239 & 15.45    & 9.239 & 15.45    & 9.239 & 15.45    & 9.239 & 15.45  & 9.239 & 15.45   \\

&\multicolumn{1}{c|}{}   & DD  & 2.191     &\textbf{\textcolor{red}{3.125}}          & 2.738         & 3.832         & 2.436          & 3.297          & 2.489       & 3.560      & 2.261           & 3.268           & \textbf{\textcolor{red}{2.180}}        & 3.154       & 2.477        & 3.390       & 2.513           & 3.384           & 2.221        & 3.275        & 2.607        & 3.401         & 2.277         & 3.201        &  \textcolor{blue}{\underline{2.181}}  &\textcolor{blue}{\underline{3.130}}   &2.222       &3.142     \\

&\multicolumn{1}{c|}{}                       & Cor & \textbf{\textcolor{red}{2.373}}       & \textbf{\textcolor{red}{3.286}}     & \rule[0.07cm]{0.8cm}{1pt}          & \rule[0.07cm]{0.8cm}{1pt}                  & 2.398           & 3.298          & 2.388       & 3.309       & 2.602           & 3.420           & 2.421        & 3.354       & 2.485        & 3.310       & 2.492           & 3.321           & \textcolor{blue}{\underline{2.380}}        & 3.312       & 2.381          & \textcolor{blue}{\underline{3.291}}     & 2.420       & 3.299            &2.472      &3.301      &2.403      &3.305                \\

\rowcolor{gray!20}
&\multicolumn{1}{c|}{}                       & Avg & \textbf{\textcolor{red}{2.282}}      & \textbf{\textcolor{red}{3.205}}       & \rule[0.07cm]{0.8cm}{1pt}         & \rule[0.07cm]{0.8cm}{1pt}               & 2.417           & 3.298          & 2.439       & 3.435       & 2.432           & 3.344           &  \textcolor{blue}{\underline{2.301}}       & 3.254       & 2.481        & 3.350       & 2.503           & 3.352           & \textcolor{blue}{\underline{2.301}}        & 3.294        & 2.494      & 3.346     &  2.349       &3.250         &   2.327        &\textcolor{blue}{\underline{3.215}}    &2.312      &3.223              \\ \cmidrule(lr){2-29}

&\multicolumn{1}{c|}{\multirow{4}{*}{\rotatebox{90}{Vis.o}}} & MM    &  125.7   & 176.4      &  125.7   & 176.4      &  125.7   & 176.4      &  125.7   & 176.4      &  125.7   & 176.4      &  125.7   & 176.4      &  125.7   & 176.4      &  125.7   & 176.4      &  125.7   & 176.4      &  125.7   & 176.4      &  125.7   & 176.4       &  125.7   & 176.4   &  125.7   & 176.4   \\

&\multicolumn{1}{c|}{}       & DD  & \textbf{\textcolor{red}{28.27}}      & \textbf{\textcolor{red}{46.84}}            & 40.80         & 61.89         & 29.56           & 47.67          & 28.75       & 47.29       & 29.43           & 48.21           & \textcolor{blue}{\underline{28.48}}       & 47.25       & 29.79        & 47.38       & 30.33           & 47.88           & 28.67        & \textcolor{blue}{\underline{46.87}}        & 30.04      & 48.48    & 28.63          & \textcolor{blue}{\underline{46.87}}       &29.49      &47.27      &29.29      &47.32          \\

&\multicolumn{1}{c|}{}                       & Cor & \textcolor{blue}{\underline{36.98}}       & \textbf{\textcolor{red}{53.48}}      & \rule[0.07cm]{0.8cm}{1pt}          & \rule[0.07cm]{0.8cm}{1pt}                  & 43.87           & 61.75          & 40.12       & 58.61       & 38.20           & 59.56           & 40.60        & 57.83       & 40.04        & 55.85       & 46.77           & 65.80           & 44.72        & 64.73        & \textbf{\textcolor{red}{36.36}}         & 56.70      & 38.98       &  \textcolor{blue}{\underline{53.95}}     &42.42      &65.33      &38.90      &57.49              \\

\rowcolor{gray!20}
&\multicolumn{1}{c|}{}                       & Avg & \textbf{\textcolor{red}{32.62}}       & \textbf{\textcolor{red}{50.16}}       & \rule[0.07cm]{0.8cm}{1pt}         & \rule[0.07cm]{0.8cm}{1pt}            & 36.72           & 54.71          & 34.44       & 52.95       & 33.82           & 53.88           & 34.54        & 52.54       & 34.92        & 51.61       & 38.55           & 56.84           & 36.69        & 55.80        & \textcolor{blue}{\underline{33.20}}          & 52.59    &33.81            & \textcolor{blue}{\underline{50.41}}            &35.96        &56.30      &34.09      &52.40      \\ \cmidrule(lr){2-29}

&\multicolumn{1}{c|}{\multirow{4}{*}{\rotatebox{90}{Vis.l}}}     & MM    &445.6   & 477.4     &445.6   & 477.4     &445.6   & 477.4     &445.6   & 477.4     &445.6   & 477.4     &445.6   & 477.4     &445.6   & 477.4     &445.6   & 477.4     &445.6   & 477.4     &445.6   & 477.4     &445.6   & 477.4     &445.6   & 477.4  &445.6   & 477.4     \\

&\multicolumn{1}{c|}{}      & DD  & \textbf{\textcolor{red}{31.82}}      &\textbf{\textcolor{red}{46.65}}          & 76.17         & 96.65         & \textcolor{blue}{\underline{32.91}}           & \textcolor{blue}{\underline{49.42}}          & 34.46       & 50.31       & 35.66           & 50.63           & 33.83        & 49.85       & 34.89        & 50.48       & 34.23           & 50.38           & 36.68        & 52.83        & 37.40             & 53.44     & 35.06         & 50.92        &33.13      &48.37      &37.59      &53.15                \\

&\multicolumn{1}{c|}{}                       & Cor & 44.00     & \textcolor{blue}{\underline{60.20}}       & \rule[0.07cm]{0.8cm}{1pt}          & \rule[0.07cm]{0.8cm}{1pt}            & 43.81           & 60.89          & 43.89       & 60.38       & 43.92           & \textcolor{blue}{\underline{60.20}}           & 44.77        & 61.88       & 44.28        & 60.27       & 43.76           & 60.35           & 43.72        & 60.29        & \textcolor{blue}{\underline{43.67}}        & 60.47         & 43.76  & 60.25        & 43.73     &  \textbf{\textcolor{red}{60.15}}     &\textbf{\textcolor{red}{43.64}}     &60.36       \\

\rowcolor{gray!20}
&\multicolumn{1}{c|}{}                       & Avg & \textbf{\textcolor{red}{37.91}}     & \textbf{\textcolor{red}{53.43}}              & \rule[0.07cm]{0.8cm}{1pt}          & \rule[0.07cm]{0.8cm}{1pt}       &  \textcolor{blue}{\underline{38.36}}          & 55.15          & 39.17      & 55.34       & 39.79           & 55.42           & 39.30        & 55.86       & 39.58        & 55.38       & 38.99           & 55.36           & 40.20        & 56.56        & 40.53       & 56.95      & 39.41         & 55.59       &38.43      &\textcolor{blue}{\underline{54.26}}        &40.62      &56.75           \\ \cmidrule(lr){2-29}

&\multicolumn{1}{c|}{\multirow{4}{*}{\rotatebox{90}{\textbf{AF}}}}     & MM    &0.951   & 2.044     &0.951   &2.044     &0.951   & 2.044     &0.951   & 2.044     &0.951   & 2.044     &0.951   & 2.044    &0.951   & 2.044    &0.951   & 2.044     &0.951   & 2.044     &0.951   & 2.044    &0.951   & 2.044  &0.951   & 2.044  &0.951   & 2.044     \\

&\multicolumn{1}{c|}{}      & DD  & \textbf{\textcolor{red}{0.087}}      &\textbf{\textcolor{red}{0.125}}          & 0.552         & 0.869         & 0.099           & 0.171         & 0.129          & 0.182         & \textcolor{blue}{\underline{0.091}}          &    \textcolor{blue}{\underline{0.152}}      & 0.102      & 0.164       & 0.110      & 0.170      & 0.188     & 0.231     & 0.093      & 0.180      & 0.120      & 0.184     & 0.145        & 0.209       &  \textcolor{blue}{\underline{0.091}}       & 0.155       & 0.098      & 0.153                 \\

&\multicolumn{1}{c|}{}                       & Cor & \textcolor{blue}{\underline{0.279}}     & 0.377       & \rule[0.07cm]{0.8cm}{1pt}          & \rule[0.07cm]{0.8cm}{1pt}            & 0.347           & 0.422          & \textbf{\textcolor{red}{0.274}}           & \textbf{\textcolor{red}{0.364}}         & 0.288          & 0.382         & 0.318        & 0.408       & 0.298      & 0.382      & 0.312      & 0.394     & 0.328      & 0.421      & 0.291      & 0.377     & 0.306        & 0.387       & 0.313      & 0.410       & 0.292     & \textcolor{blue}{\underline{0.375}}           \\

\rowcolor{gray!20}
&\multicolumn{1}{c|}{}                       & Avg & \textbf{\textcolor{red}{0.183}}     & \textbf{\textcolor{red}{0.251}}              & \rule[0.07cm]{0.8cm}{1pt}          & \rule[0.07cm]{0.8cm}{1pt}       &  0.223          & 0.296          & 0.201         & 0.273        &\textcolor{blue}{\underline{0.189}}          & 0.267         & 0.210        & 0.286       & 0.204      & 0.276      & 0.250      & 0.312     & 0.211      & 0.301      & 0.205      & 0.280     & 0.226        & 0.298       & 0.202       & 0.283       & 0.195      & \textcolor{blue}{\underline{0.264}} \\

\addlinespace[3pt] \midrule \addlinespace[3pt] \midrule  \addlinespace[3pt]

\multicolumn{1}{c|}{\multirow{4}{*}{\rotatebox{90}{\textbf{Stage \Rmnum{2}}}}}

& \multicolumn{1}{c|}{\multirow{2}{*}{\rotatebox{90}{HL}}}     & MM   & 5.417 &   7.199   & 5.417 &   7.199   & 5.417 &   7.199   & 5.417 &   7.199   & 5.417 &   7.199   & 5.417 &   7.199   & 5.417 &   7.199   & 5.417 &   7.199   & 5.417 &   7.199   & 5.417 &   7.199   & 5.417 &   7.199   & 5.417 &   7.199  & 5.417 &   7.199  \\

&\multicolumn{1}{c|}{} & DD & \textbf{\textcolor{red}{1.556}}       & \textbf{\textcolor{red}{2.677}}     & 1.602         & 2.679         & 1.758           & 2.905          & 1.728       & 2.789       & 1.651           & 2.757           & \textcolor{blue}{\underline{1.574}}        & \textcolor{blue}{\underline{2.698}}       & 1.852        & 2.871       & 1.644           & 2.805           & 1.670        & 2.813        & 1.622 & 2.734    & 1.752 & 2.818      &1.894    &2.969  &1.647  &2.765           \\ \cmidrule(lr){2-29}

&\multicolumn{1}{c|}{\multirow{2}{*}{\rotatebox{90}{P}}} & MM

& 77.55  & 131.2& 77.55  & 131.2& 77.55  & 131.2& 77.55  & 131.2& 77.55  & 131.2& 77.55  & 131.2& 77.55  & 131.2& 77.55  & 131.2& 77.55  & 131.2& 77.55  & 131.2& 77.55  & 131.2& 77.55  & 131.2     &77.55  & 131.2\\

&\multicolumn{1}{c|}{} &DD  & \textbf{\textcolor{red}{3.017}}      & \textbf{\textcolor{red}{5.424}}         & 14.60         & 29.03         & 9.067           & 13.73          & 7.471       & 10.12       & 10.41           & 22.39           &  \textcolor{blue}{\underline{3.527}}       & \textcolor{blue}{\underline{6.234}}       & 4.488        & 7.173       & 4.695           & 7.881           & 9.281        & 17.39        & 4.426 & 7.390    & 9.885 & 14.69      &33.51      &69.99      &20.51      &41.01               \\

\specialrule{3pt}{3pt}{0pt}

\end{tabular}%

}}

\label{table1}

\end{table*}

The overall performance is summarized in Tab.\ref{table1}. In Stage \Rmnum{1}, under the DNA architecture, each model is required not only to predict the target variable but also to produce a correction function to further refine the mechanistic model output. Therefore, we evaluate the predicted target values (DD), the corrected mechanistic model outputs (Cor), and their average (Avg). In Stage \Rmnum{2}, since models are only required to predict the target variable, we evaluate predictive performance only (DD). Furthermore, we include the mechanistic model (MM) results to highlight the improvements achieved by different models under the DNA architecture. \textbf{Stage \Rmnum{1}.} First, in the vertical comparison in the table, we observe that for almost all baselines, both the DD and Cor values show significant improvement compared to MM: the improvement in DD indicates that these baseline models possess strong predictive capabilities, validating the rationality of our baseline selection; the substantial gains in Cor can be attributed to the DNA architecture, under which data-driven models provide effective corrections to the mechanistic model. Regardless of which data-driven model is employed within the DNA architecture, the architecture consistently brings the mechanistic outputs to a substantially higher level of fidelity. Specifically, on approximately 44\% of the datasets (Cp.g, Cp.o, Rho.o, Vis.l), Cor achieves an order-of-magnitude improvement over MM in terms of MAE. On the remaining 56\% (Cp.w, Rho.l, Vis.w, Vis.o, AF), the MAE is reduced by a factor of 2 to 4. Second, in the horizontal comparison in the table, we find that our proposed data-driven framework consistently achieves the best Avg performance on at least one metric for every dataset. On 88.9\% of the datasets (8 out of 9), it outperforms all baselines across all metrics, demonstrating the model’s stability and strong generalization ability. In particular, our model achieves the best DD performance on at least one metric in 88.9\% (8 out of 9) of the datasets, further demonstrating the superior predictive capability of the proposed data-driven framework in Stage I. \textbf{Stage \Rmnum{2}.} In this phase, we evaluate only the predicted target values of each model. DNA achieves the best results across all metrics on both datasets. Specifically, compared to the strongest baseline MultiPatchFormer \cite{naghashi2025multiscale}, DNA reduces MAE and RMSE on these two datasets by 7.80\% and 6.89\% on average, respectively.


\subsection{Ablation Study}

\begin{wraptable}{r}{0.42\textwidth}
\centering
\caption{Ablation studys on Stage \Rmnum{2}.}

\resizebox{0.4\textwidth}{!}{%
\begin{tabular}{ccc|cc|cc}
\specialrule{1.2pt}{0pt}{1.5pt}
\multicolumn{3}{c|}{\multirow{2}{*}[-0.5ex]{Stage \Rmnum{2}}}
& \multicolumn{2}{c|}{HL} & \multicolumn{2}{c}{P} \\
\cmidrule(lr){4-7}

\multicolumn{3}{c|}{}
& MAE & RMSE & MAE & RMSE \\
\specialrule{1.2pt}{1.5pt}{1.5pt}

\multicolumn{3}{c|}{w/o GCN-E}
& 1.612 & 2.726 & 4.509 & 7.641 \\

\multicolumn{3}{c|}{w/o Gating}
& 1.757 & 2.891 & 8.858 & 12.48 \\

\multicolumn{3}{c|}{\textbf{Ours}}
& \textbf{\textcolor{red}{1.556}} & \textbf{\textcolor{red}{2.677}}
& \textbf{\textcolor{red}{3.017}} & \textbf{\textcolor{red}{5.424}} \\

\specialrule{1.2pt}{1.5pt}{0pt}
\end{tabular}%
}
\label{table4}
\end{wraptable}

\textbf{Stage \Rmnum{1}.} The ablation study results for Stage \Rmnum{1} are summarized in Tab.\ref{table2}. To verify the superiority of the proposed MASA over the vanilla multi-head self-attention (MHSA), we first replace MASA with MHSA, and the experimental results (MASA $\rightarrow$ MHSA) show that replacing MASA with MHSA leads to a clear performance degradation. Specifically, the average MAE and RMSE across the eight datasets increase by 10.21\% and 7.92\%, respectively. In addition, we further evaluate the effectiveness of each component by removing them individually, including the Tri-Embedding (Tri-E), MASA, and the channel attention with ResBlocks (CA\&RB). The results indicate that the full combination of Tri-E, MASA, and CA\&RB achieves the most accurate predictions on nearly all datasets, while removing any single component consistently leads to degraded performance.

\begin{table*}[]

\caption{Ablation study results on Stage \Rmnum{1}.}
\centering

\resizebox{\textwidth}{!}{%

\begin{tabular}{cccc|cc|cc|cc|cc|cc|cc|cc|cc}

\specialrule{1.2pt}{0pt}{1.5pt}

\multicolumn{4}{c|}{\multirow{2}{*}[-0.5ex]{Stage \Rmnum{1}}}                       & \multicolumn{2}{c|}{Cp.g} & \multicolumn{2}{c|}{Cp.o} & \multicolumn{2}{c|}{Cp.w} & \multicolumn{2}{c|}{Rho.o} & \multicolumn{2}{c|}{Rho.l} & \multicolumn{2}{c|}{Vis.w} & \multicolumn{2}{c|}{Vis.o} & \multicolumn{2}{c}{Vis.l} \\  \cmidrule(lr){5-20}

\multicolumn{4}{c|}{}    & MAE         & RMSE        & MAE         & RMSE        & MAE         & RMSE        & MAE          & RMSE        & MAE          & RMSE        & MAE          & RMSE        & MAE          & RMSE        & MAE         & RMSE        \\ \specialrule{1.2pt}{1.5pt}{1.5pt}

\multicolumn{3}{c|}{\multirow{3}{*}{\textbf{Ours}}} & DD  & \textbf{\textcolor{red}{0.763}}      & \textbf{\textcolor{red}{1.073}}       & {1.704 }      & \textbf{\textcolor{red}{2.335}  }     & \textbf{\textcolor{red}{0.724} }    & \textbf{\textcolor{red}{1.017}}       & \textbf{\textcolor{red}{0.793}}        & \textbf{\textcolor{red}{1.367}}       &{2.434}       & {3.437}       & 2.191       & \textbf{\textcolor{red}{3.125}  }     & \textbf{\textcolor{red}{28.27}}        & 46.84       & {31.82}      & \textbf{\textcolor{red}{46.65}}      \\

\multicolumn{1}{c}{}   & \multicolumn{1}{c}{}  &   \multicolumn{1}{c|}{}                  & Cor & 0.831       & \textbf{\textcolor{red}{1.067}}       & 1.895       & \textbf{\textcolor{red}{2.474} }      & {1.219}       & {1.611}       & \textbf{\textcolor{red}{1.504} }       &{ 2.397 }      & \textbf{\textcolor{red}{6.583}  }      & \textbf{\textcolor{red}{7.946 }}      & \textbf{\textcolor{red}{2.373}}       &\textbf{ \textcolor{red}{3.286 }}      & \textbf{\textcolor{red}{36.98}}        & {53.48}      & {44.00}       & \textbf{\textcolor{red}{60.20}}      \\

\rowcolor{gray!20}
\multicolumn{1}{c}{}   & \multicolumn{1}{c}{}  &   \multicolumn{1}{c|}{}              & Avg & \textbf{\textcolor{red}{0.797}}       & \textbf{\textcolor{red}{1.070}}     & \textbf{\textcolor{red}{1.799}}      & \textbf{\textcolor{red}{2.404}}     & \textbf{\textcolor{red}{0.972}  }    & \textbf{\textcolor{red}{1.314 }}      & \textbf{\textcolor{red}{1.148} }      & \textbf{\textcolor{red}{1.882}}     & {4.509}        &\textbf{\textcolor{red}{ 5.692}  }    & \textbf{\textcolor{red}{2.282}}        & \textbf{\textcolor{red}{3.205} }      & \textbf{\textcolor{red}{32.62}  }     & \textbf{\textcolor{red}{50.16}  }     & \textbf{\textcolor{red}{37.91}}      &\textbf{\textcolor{red}{53.43} }     \\ \midrule

\multicolumn{3}{c|}{\multirow{3}{*}{MASA $\rightarrow$ MHSA}}  & DD  & 0.963       & 1.231       & \textbf{\textcolor{red}{1.702} }      & 2.372       & 1.333       & 1.532       & 0.802        & 1.394       & \textbf{\textcolor{red}{2.144} }       & 3.451       & 2.252        & 3.179       & 28.31        & 46.82      & 32.10      & 47.19       \\

\multicolumn{1}{c}{}   & \multicolumn{1}{c}{}  &   \multicolumn{1}{c|}{}               & Cor & \textbf{\textcolor{red}{0.787 }}      & 1.091       & 1.962      & 2.543       & {1.167 }      & {1.542 }      & 1.575        & \textbf{\textcolor{red}{2.382}}      & 6.726       & 8.151        & 2.520       & 3.311       & 59.74        & 85.06       & \textbf{\textcolor{red}{43.72}}      & 60.46      \\

\rowcolor{gray!20}
\multicolumn{1}{c}{}   & \multicolumn{1}{c}{}  &   \multicolumn{1}{c|}{}                 & Avg & 0.875       & 1.161      & 1.832      & 2.457       & 1.250       & 1.537       & 1.189       & 1.888       &\textbf{\textcolor{red}{ 4.435}}        & 5.801       & 2.386       & 3.245       & 44.03       & 65.94       & \textbf{\textcolor{red}{37.91} }     & 53.82       \\ \midrule

\multicolumn{3}{c|}{\multirow{3}{*}{w/o Tri-E}}  & DD  & 1.579       & 1.848       & 2.045       & 2.720       & 2.971       & 3.566      & 1.753        & 2.202       & 2.302        & \textbf{\textcolor{red}{3.359}}       & 2.221        & 3.173       & 28.47        & 46.81       & \textbf{\textcolor{red}{31.43} }      & 47.43       \\

\multicolumn{1}{c}{}   & \multicolumn{1}{c}{}  &   \multicolumn{1}{c|}{}               & Cor & 0.883       & 1.130       & 1.910       & 2.491       & \textbf{\textcolor{red}{1.113} }      &   \textbf{\textcolor{red}{1.457}}     & 1.724       & 2.424       & 6.741        & {8.115}        & {2.770}        & {3.678}       & 39.61        & 58.03       & 45.73     & 62.89      \\

\rowcolor{gray!20}
\multicolumn{1}{c}{}   & \multicolumn{1}{c}{}  &   \multicolumn{1}{c|}{}                 & Avg & 1.231       & 1.489       & 1.977      & 2.605       &2.042     & 2.512      & 1.738        & {2.313}       & 4.522        & 5.737      & 2.495        & 3.426       & 34.04        & 52.42       & 38.58      & 55.16     \\ \midrule

\multicolumn{3}{c|}{\multirow{3}{*}{w/o MASA}} & DD  & 0.989       & 1.297       & 1.777       & 2.483       & 1.567       & 1.847       & 0.928        & 1.420      & 2.171        & 3.368       & {2.219}        & 3.212       & 28.49        & \textbf{\textcolor{red}{46.80} }      & 32.11       & 47.70       \\

\multicolumn{1}{c}{}   & \multicolumn{1}{c}{}  &   \multicolumn{1}{c|}{}                & Cor & {0.846}       &{ 1.226}       & \textbf{\textcolor{red}{1.876}}       & {2.475}      & 1.615       & 2.010      & 1.559        & \textbf{\textcolor{red}{2.382}   }    & {6.794}        & 8.243       & 2.513       & 3.315       & 38.58        & 55.10       & 44.97       & 61.88       \\

\rowcolor{gray!20}
\multicolumn{1}{c}{}   & \multicolumn{1}{c}{}  &   \multicolumn{1}{c|}{}                    & Avg & 0.917       & 1.262       & 1.827       & 2.479      & 1.591       & 1.929      & 1.244        & 1.901       & 4.483       & 5.806       & 2.366        & 3.263       & 33.53        & 50.95       & 38.54      & 54.79       \\ \midrule

\multicolumn{3}{c|}{\multirow{3}{*}{w/o CA\&RB}} & DD  & 1.314       & 1.641       & 1.771       & 2.425       & 3.139       & 3.566      & 1.088        & 1.665       & 2.481        & 3.750       & \textbf{\textcolor{red}{2.177} }      & 3.138      & 29.00       & 47.41       & 37.00      & 51.66       \\

\multicolumn{1}{c}{}   & \multicolumn{1}{c}{}  &   \multicolumn{1}{c|}{}                  & Cor & 0.886       & 1.133       & 2.029       & 2.555       & 1.487       & 1.985       & 1.623       & 2.386       & 6.889        & 8.391      & 2.511        & 3.330      & 37.74       & \textbf{\textcolor{red}{53.32}}      & 43.79      & 60.48      \\

\rowcolor{gray!20}
\multicolumn{1}{c}{}   & \multicolumn{1}{c}{}  &   \multicolumn{1}{c|}{}              & Avg & 1.100      & 1.387       & 1.900      & 2.490       & 2.313       & 2.775      & 1.356       & 2.026       & 4.685        & 6.070       & 2.344        & 3.234       & 33.37        & 50.36       & 40.40      & 56.07       \\

\specialrule{1.2pt}{1.5pt}{0pt} 

\end{tabular}%

}

\label{table2}
\end{table*}

\begin{table}[]
\caption{Comparison of Stage II Outputs of the Original and Corrected Mechanistic Models.}
\label{tab3}
\Huge
\resizebox{\textwidth}{!}{%
\begin{tabular}{ccc|cc|cc|cc|cc|cc|cc|cc|cc|cc|cc|cc|cc}
\toprule
\multicolumn{3}{c|}{\textbf{Model}}  & \multicolumn{2}{c|}{\textbf{DNA}} & \multicolumn{2}{c|}{Transformer} & \multicolumn{2}{c|}{FlowformerMH} & \multicolumn{2}{c|}{iTransformer} & \multicolumn{2}{c|}{MPFormer} & \multicolumn{2}{c|}{GATv2} & \multicolumn{2}{c|}{DGDL} & \multicolumn{2}{c|}{TGCNS} & \multicolumn{2}{c|}{RaDA} & \multicolumn{2}{c|}{TimesNet} & \multicolumn{2}{c|}{TimeXer} & \multicolumn{2}{c|}{Umixer} \\ \midrule
\multicolumn{3}{c|}{\textbf{Metric}} & MAE             & RMSE       & MAE             & RMSE     & MAE             & RMSE           & MAE             & RMSE           & MAE           & RMSE         & MAE         & RMSE        & MAE         & RMSE       & MAE         & RMSE        & MAE         & RMSE       & MAE           & RMSE         & MAE          & RMSE         & MAE         & RMSE        \\\midrule

& \multicolumn{1}{c|}{\multirow{2}{*}{\rotatebox{90}{HL}}}     &MM  & 8.513 &   10.91  & 8.513 &   10.91& 8.513 &   10.91& 8.513 &   10.91& 8.513 &   10.91& 8.513 &   10.91& 8.513 &   10.91& 8.513 &   10.91& 8.513 &   10.91& 8.513 &   10.91& 8.513 &   10.91& 8.513 &   10.91 \\

&\multicolumn{1}{c|}{} & Cor & 5.417      & 7.199         & 5.445           & 7.175          & 5.449       & 7.180       & 5.424           & 7.129           & 5.418        & 7.129       & 5.419        & 7.133       & \textcolor{blue}{\underline{5.414}}          & 7.131           & 5.416        & 7.136       & \textbf{\textcolor{red}{5.413}}  & \textcolor{blue}{\underline{7.126}}    & 5.426  & 7.142      &5.419    &\textbf{\textcolor{red}{7.117}}  &5.440 &7.148           \\ 

\cmidrule(lr){2-27}

& \multicolumn{1}{c|}{\multirow{2}{*}{\rotatebox{90}{P}}}     &MM  & 124.2 &   214.5  & 124.2 &   214.5 & 124.2 &   214.5 & 124.2 &   214.5 & 124.2 &   214.5 & 124.2 &   214.5 & 124.2 &   214.5 & 124.2 &   214.5 & 124.2 &   214.5 & 124.2 &   214.5 & 124.2 &   214.5 & 124.2 &   214.5  \\

&\multicolumn{1}{c|}{} & Cor & \textbf{\textcolor{red}{77.55}}       & \textbf{\textcolor{red}{131.2}}         & 92.05          & 180.1         & \textcolor{blue}{\underline{90.22}}       & \textcolor{blue}{\underline{171.4} }      & 93.49           & 184.8           & 90.47       & 177.1       & 92.03        & 179.4      & 91.91           & 179.3           & 91.83        & 179.3        & 91.69    & 178.9    & 90.90  & 177.6      &94.94    &187.0  &90.62  &176.7           \\

\bottomrule

\end{tabular}%
}
\end{table}

\textbf{Stage \Rmnum{2}.} The ablation study results for Stage II are summarized in Tab.\ref{table4}. These experimental results show that the model incorporating both the GCN-Embedding (GCN-E) and the gating mechanism (Gating) achieves the best performance. In contrast, removing either component results in varying degrees of performance degradation, demonstrating the effectiveness and necessity of each module.


\subsection{Analysis of Correction Function}

The introduction of a unique interaction strategy, in which the correction function $\mathcal{F}$ is generated and fed back to the mechanistic model to improve its performance is one of the key contributions of this work. The correction functions considered in this work consist of three components: (1) an $n$ order polynomial term with  $n \in \{0, 1, 2, 3\}$; (2) a logarithmic term (with or without $ln$); and (3) weighting coefficients (with or without $gate$). We conducted extensive experiments to select the most suitable combination of the three components for each Stage I dataset, thereby determining a predefined function form. Under the DNA architecture, all baseline models were trained using the same predefined function forms to ensure a fair comparison. The predefined function form for each dataset and the correction functions obtained by our model are summarized in Tab.\ref{table6}.


\subsection{Analysis of Gain Propagation Along the DNA Chains}

\begin{figure}
    \centering
    \includegraphics[width=\linewidth]{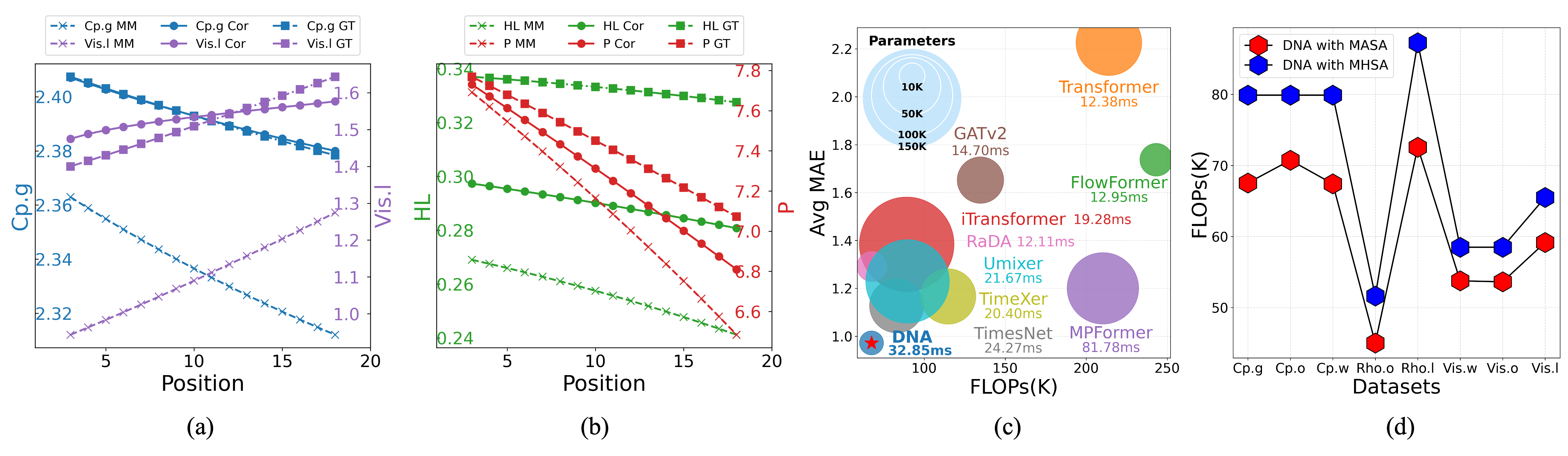}
    \caption{(a,b) Visualization and efficiency comparison results. (c) Model complexity and efficiency comparison on the Cp.w dataset. (d) Computational cost comparison between MASA and MHSA across all Stage I datasets of the three-phase pipeline system}
    \label{fig3}
\end{figure}

We deploy the corrected mechanistic model $\mathcal{M}^{I}_{cor}$ obtained by applying DNA with each baseline in Stage I (green pipeline in Fig.\ref{fig1}), keeping everything else unchanged, and repeat the experiment under identical operating conditions $X$, with one run for each baseline. The resulting Stage II outputs are reported in Tab.\ref{tab3}, where MM denotes the outputs of the original mechanistic model and Cor those of the corrected model. As shown in Tab.\ref{table1}, under the DNA architecture, every baseline could substantially improve the predictive performance of $\mathcal{M}^{I}_{{me}}$, yielding $\mathcal{M}^{I}_{{cor}}$. Combined with Tab.\ref{tab3}, when these $\mathcal{M}^{I}_{{cor}}$ models are deployed in Stage I, the final outputs are markedly improved even with the downstream model ($\mathcal{M}^{II}_{{me}}$) unchanged. This shows that, under DNA architecture, the gains achieved upstream can be effectively propagated to downstream tasks. Specifically, our model achieves average MAE reductions of 36.37\% and 37.56\% for HL and P, respectively, and achieves the best performance on P, with a further 14.04\% improvement in MAE over the second-best model. To intuitively demonstrate the improvement brought by this correction, we also visualize the results on 16 positions from a randomly selected case, as shown in Fig.\ref{fig3}a-b.

\subsection{Analysis of Model Complexity and Efficiency}

\textbf{Complexity.} We compare the complexity of our model and other baselines in terms of total number of parameters and FLOPs required for a single forward inference. As shown in Fig.\ref{fig3}c, under identical configurations, our model achieves the best predictive performance while also having the lowest total number of parameters and FLOPs among all models. This demonstrates that our model is highly lightweight. We further demonstrate the effectiveness of the proposed MASA in reducing computational cost, as shown in Fig.\ref{fig3}d. Under strictly identical configurations, replacing MHSA with MASA leads to a clear reduction in the overall FLOPs across all Stage \Rmnum{1} datasets of the three-phase pipeline system, with an average reduction of 12.28\%. \textbf{Efficiency.} We also evaluate efficiency by comparing the inference time required by different models to process a batch of samples. As shown in Fig.\ref{fig3}c, although the experimental results indicate that DNA does not exhibit superior inference speed among these baselines, its favorable FLOPs performance suggests that DNA has the potential for efficient inference in theory. After careful investigation and analysis, we attribute this to the lack of an effective strategy for parallelizing MASA, which currently results in suboptimal inference efficiency. More experimental results and analyses are provided in the Appendix \ref{more results}.

\section{Conclusion, Future Work and Limitations}

In this work, we introduced a novel Double-Helix Hierarchical Deep Interaction Architecture (DNA) for industrial digital twin. The main idea is to enable effective interaction between mechanistic and data-driven models, ultimately building accurate models of complex industrial systems. A key objective of this work is to leverage data-driven models not only for accurate predictions, but also to assist in improving the performance of mechanistic models. This enhancement is of great practical and industrial value. We validate DNA through extensive experiments on the three-phase pipeline and acentric factor systems, comparing it with many competitive models. Results show DNA consistently outperforms baselines, and we believe DNA has the potential to contribute meaningfully to the advancement of industrial digital twins.


Other directions for future work are, in part, motivated by the limitations of this work: (1) DNA has the potential for efficient inference, but this potential remains unrealized. With the underlying cause identified, exploring how to parallelize MASA is a natural next step. (2) Currently, the forms of the correction functions are carefully designed in advance, and their end-to-end generation remains to be investigated. (3) DNA is limited to correcting mechanistic models formulated as explicit functions. We plan to extend it to mechanistic models governed by systems of differential equations. (4) Finally, we leave the exploration of applications in broader industrial scenarios to future work.




\bibliography{references}
\bibliographystyle{unsrtnat}

\appendix

\section*{Supplemental Material}
This supplemental material is divided into the following five appendices:

\begin{itemize}
    \item \textbf{Appendix A}: Related works on existing structured hybrid modeling.
    \item \textbf{Appendix B}: Additional details of the proposed data-driven modeling frameworks in Stages I and II in the Methodology section.
    \item \textbf{Appendix C}: Details of the three-phase pipeline system, including an introduction to three-phase pipeline, dataset acquisition, mechanistic models, and the data flow within the system.
    \item \textbf{Appendix D}: Details of the acentric factor system, including an introduction to acentric factor, dataset acquisition and the mechanistic models.
    \item \textbf{Appendix E}: Additional details on the Experiment section.
\end{itemize}

\begin{figure}
\centering
\includegraphics[width=0.8\textwidth]{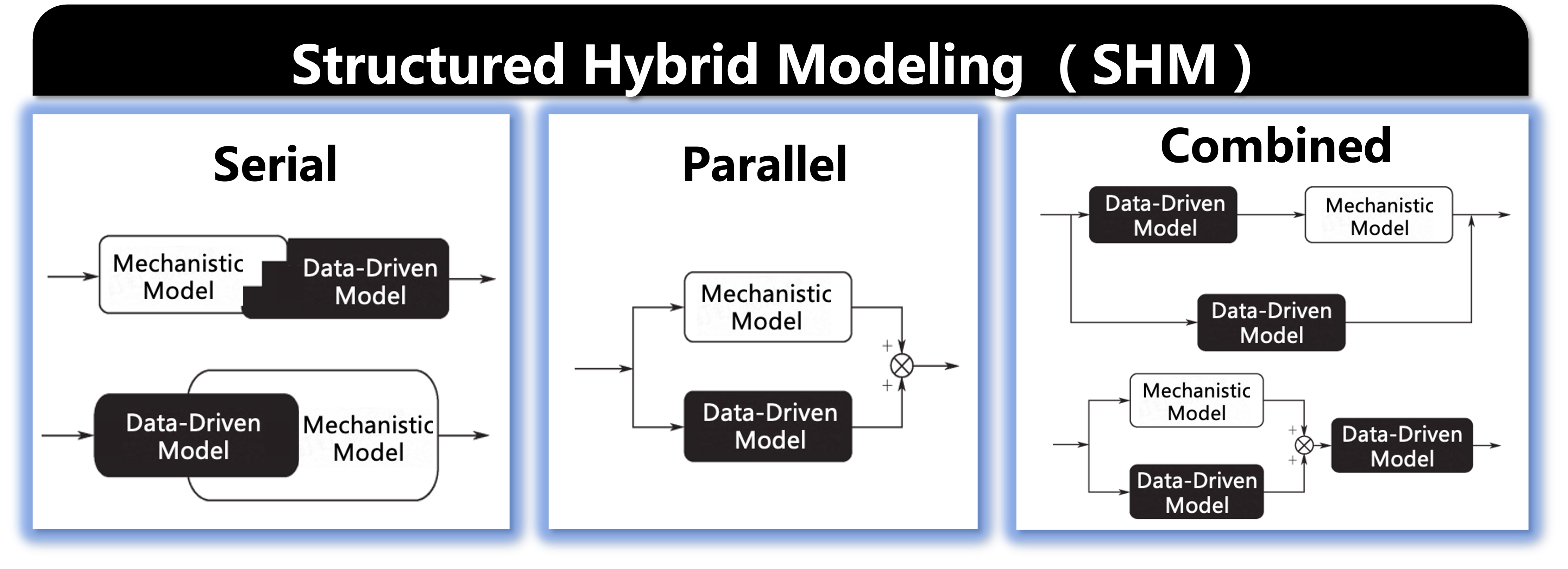}
\caption{Three common paradigms of structured hybrid modeling.}
\label{fig4-}
\end{figure}

\section{Related Work}

Industrial modeling can be broadly categorized into three paradigms: mechanistic modeling, data-driven modeling, and hybrid modeling \cite{glassey2018hybrid,fisher2020considerations,lin2025advancing}. Here, we focus on hybrid modeling methods. 

\textbf{Serial architecture.} This architecture is particularly useful when a certain dynamics is not accurately described by the mechanistic model. For example, Zhang et al. \cite{zhang2015real} developed a BP ANN to estimate unknown kinetic reaction rates in a mechanistic model, and then fed the predicted results into a subsequent mechanistic component to construct a hybrid model for the gold cyanidation leaching process. Similarly, Dalmau et al. \cite{dalmau2015comparison} modeled the fouling process in a membrane bioreactor using a tree-based model, which was subsequently integrated into a mechanistic framework. Gao et al. \cite{gao2026hybrid} first used an MLP to estimate a subset of process parameters in the mechanistic model, and then fed these estimates into a downstream mechanistic component for computation.

\textbf{Parallel architecture.} This architecture is typically used to model the residual error of mechanistic models in a post-hoc correction manner. For instance, Gao et al. \cite{gao2026mechanistic} developed an accurate model for multi-material composite columns by first using XGBoost to predict the residual between mechanistic predictions and ground truth, and then adding the estimated residual to the theoretical output of the mechanistic model. Similarly, Ghosh et al. \cite{ghosh2021model} designed a data-driven residual model to estimate the error of a mechanistic model, which was then added to its output to enable model predictive control in a seeded batch crystallization process. 

\textbf{Combined architecture.} Combined architecture integrates both serial and parallel coupling strategies, thus combining their respective advantages. For example, in the Neural ODE model proposed by Quaghebeur \cite{quaghebeur2022hybrid} for a water resource recovery facility, the data-driven component is first coupled with the mechanistic component in a parallel manner, forming an intermediate hybrid module, which is then integrated into a subsequent serial computational structure. Chen et al. \cite{chen2020framework} systematically explored serial, parallel, and hybrid coupling strategies for modeling a continuous stirred-tank reactor (CSTR) and two pharmaceutical case studies. By comparing the predictive performance of these three hybrid paradigms, they demonstrate that the hybrid structure achieves superior robustness against disturbances as well as better extrapolation capability.

Although these hybrid modeling approaches can effectively integrate mechanistic models with data-driven counterparts and leverage the strengths of both components, they largely rely on a simple stacking of the mechanistic and data-driven models where information flows in a unidirectional manner without enabling bidirectional interaction. Moreover, their one-shot modeling strategies often become inadequate when dealing with highly complex industrial systems.

\section{Methodology}

\subsection{Stage \Rmnum{1}}
\subsubsection{Tri-Embedding} 
\label{Tri-Embedding}

In this paper, we propose a novel embedding method, tripartite-embedding (Tri-Embedding), which generates high-dimensional representations for low-dimensional variables. This method is particularly suitable for scenarios where the mechanistic model underlying the target system exhibits a complex topological structure, as its embedding generation involves only a few nonlinear projections, resulting in minimal computational overhead. The Tri-Embedding process consists of three components, each detailed as follows:

\begin{itemize}
    \item[(\romannumeral1)] \textbf{Variable Type Encoding ($\mathcal{A}_1$).} We first assign a real-valued type token $v$ to each variable according to its physical meaning. Variables sharing the same physical meaning receive the same token. For example, in the $Cp.o$ dataset, $x_2$ denotes the boiling point ($Tb$) of the substance and $x_3$ represents the actual temperature ($T$); both belong to the temperature category and are therefore assigned the same value. In contrast, variables of different types, such as $x_2$ for the boiling point ($Tb$) and $x_1$ for the specific gravity ($SG$), are assigned distinct values. Subsequently, these type tokens are passed through an embedding layer ($EL$) followed by layer normalization ($LN$) to produce the first representation, $e^{(1)}_i = LN(EL(v))$. 
    \item[(\romannumeral2)] \textbf{Nonlinear Projection ($\mathcal{A}_2$).} To enrich the representational capacity of each variable, we apply a nonlinear projection that maps the original one-dimensional real value $x_i$ or $z_i$ into a higher-dimensional latent space. This high-dimensional embedding is then processed with layer normalization to yield the second representation, $e^{(2)}_i = LN(\sigma(Linear(x_i)))$. 
    \item[(\romannumeral3)] \textbf{Variable Context Encoding ($\mathcal{A}_3$).} We abstract the contextual information of each variable within the topological graph as a 4-element array, $context\_{info}:[level,IO,in\_degree,out\_degree]$, where $level$ denotes the depth level of the variable in the graph, $IO$ indicates the variable type \{input(1), output(2), intermediate(0), or both input and output(3)\}, $in\_degree$ and $out\_degree$ represent the in-degree and out-degree of the node, respectively. Each of the four elements is independently embedded using separate embedding layers followed by layer normalization. The resulting vectors are concatenated to form the third part of the representation, $e^{(3)}_i = CAT(LN(EL(context\_info)))$.

\end{itemize}

After obtaining the output of each component, we concatenate all three parts to generate the full high-dimensional representation of the variable $i$, $e_i = CAT(e^{(1)}_i ,e^{(2)}_i ,e^{(3)}_i )$.

\subsubsection{ResBlocks}
\label{res}

\begin{figure}
    \centering
    \includegraphics[width=0.6\linewidth]{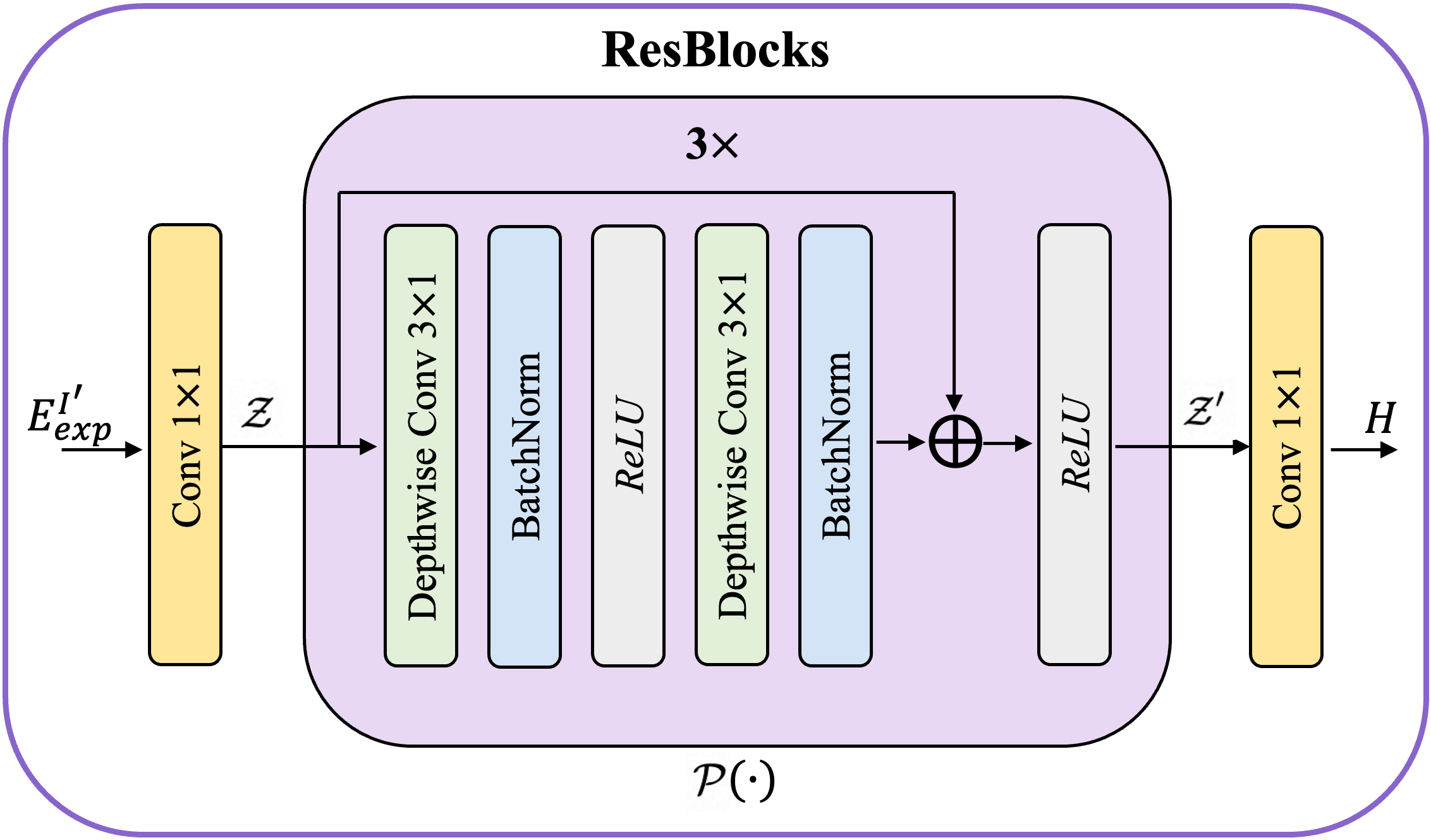}
    \caption{Diagram of ResBlocks.}
    \label{fig6}
\end{figure}

The design of the ResBlocks is illustrated in Fig.\ref{fig6}. We first apply a $1 \times 1$ convolution to the input $E_{exp}^{I^\prime}\in \mathbb{R}^{n\times d_m\times 1}$ to expand the channel dimension, yielding an intermediate representation $\mathcal{Z}\in \mathbb{R}^{c\times d_m\times 1}$. The resulting features are then processed by three consecutive ResBlocks ($\mathcal{P}(\cdot)$), producing the output $\mathcal{Z}^\prime$. Finally, another $1 \times 1$ convolution is applied to project the channel dimension back to that of the input, resulting in the final output $H$.

Notably, we employ depthwise convolutions instead of standard convolutions within each ResBlock to reduce the number of model parameters and the computational cost. Experimental results indicate that this replacement leads to no noticeable degradation in model performance. The overall process can be formulated as follows:

\begin{equation}
    \mathcal{Z}=Conv_{1\times1}(E_{exp}^{I^\prime})\in \mathbb{R}^{c\times d_m\times 1}, n \to c,
    \label{a-equa4}
\end{equation}
\begin{equation}
    \mathcal{Z}^\prime=\mathcal{P}(\mathcal{Z})\in \mathbb{R}^{c\times d_m\times 1}, c\to c,
\end{equation}
\begin{equation}
    H = Conv_{1\times1}(\mathcal{Z}^\prime)\in \mathbb{R}^{n\times d_m\times 1}, c\to n,
    \label{a-equa6}
\end{equation}

\noindent where $c$ denotes the number of channels after expansion, and we set it to 32 in this work. If we denote the entire process by $\mathcal{R}(\cdot)$, then Eqs. \ref{a-equa4}-\ref{a-equa6} correspond exactly to Eq. \ref{equa-2}  in the main text.

\subsubsection{Channel Attention Module}
\label{Channel}

\begin{figure}
    \centering
    \includegraphics[width=0.5\linewidth]{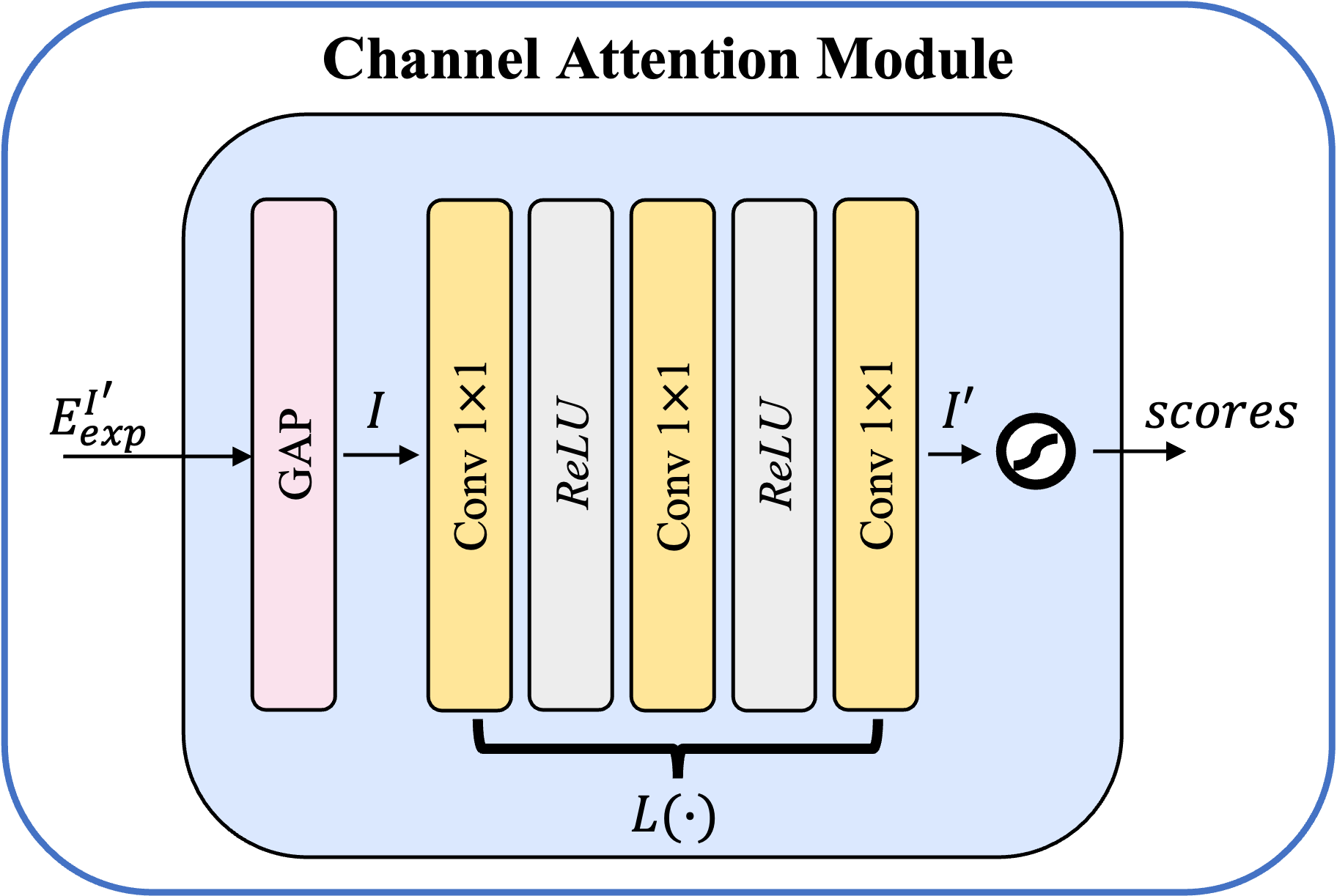}
    \caption{Diagram of Channel Attention Module.}
    \label{fig5}
\end{figure}

The channel attention module is illustrated in Fig.\ref{fig5}, and its calculation process is described as follows:

\begin{equation}
    I = GAP(E_{exp}^{I^\prime},dim=(-2,-1))\in \mathbb{R}^{n\times1\times1},
    \label{a-equa1}
\end{equation}
\begin{equation}
    I^\prime = {L}(I)\in \mathbb{R}^{n\times1\times1},n\to d_m\to \frac{d_m}{2} \to n,
\end{equation}
\begin{equation}
    scores = Sigmoid(I^\prime)\in \mathbb{R}^{n\times1\times1},
    \label{a-equa3}
\end{equation}

\noindent where $n$ is the total number of variables in the input, and $d_m$ is the embedding dimension. The steps described in Eqs. \ref{a-equa1}-\ref{a-equa3} correspond exactly to those in Eq. \ref{equa-7} of the main paper.

\subsection{Stage \Rmnum{2}}

\begin{figure}
    \centering
    \includegraphics[width=0.8\linewidth]{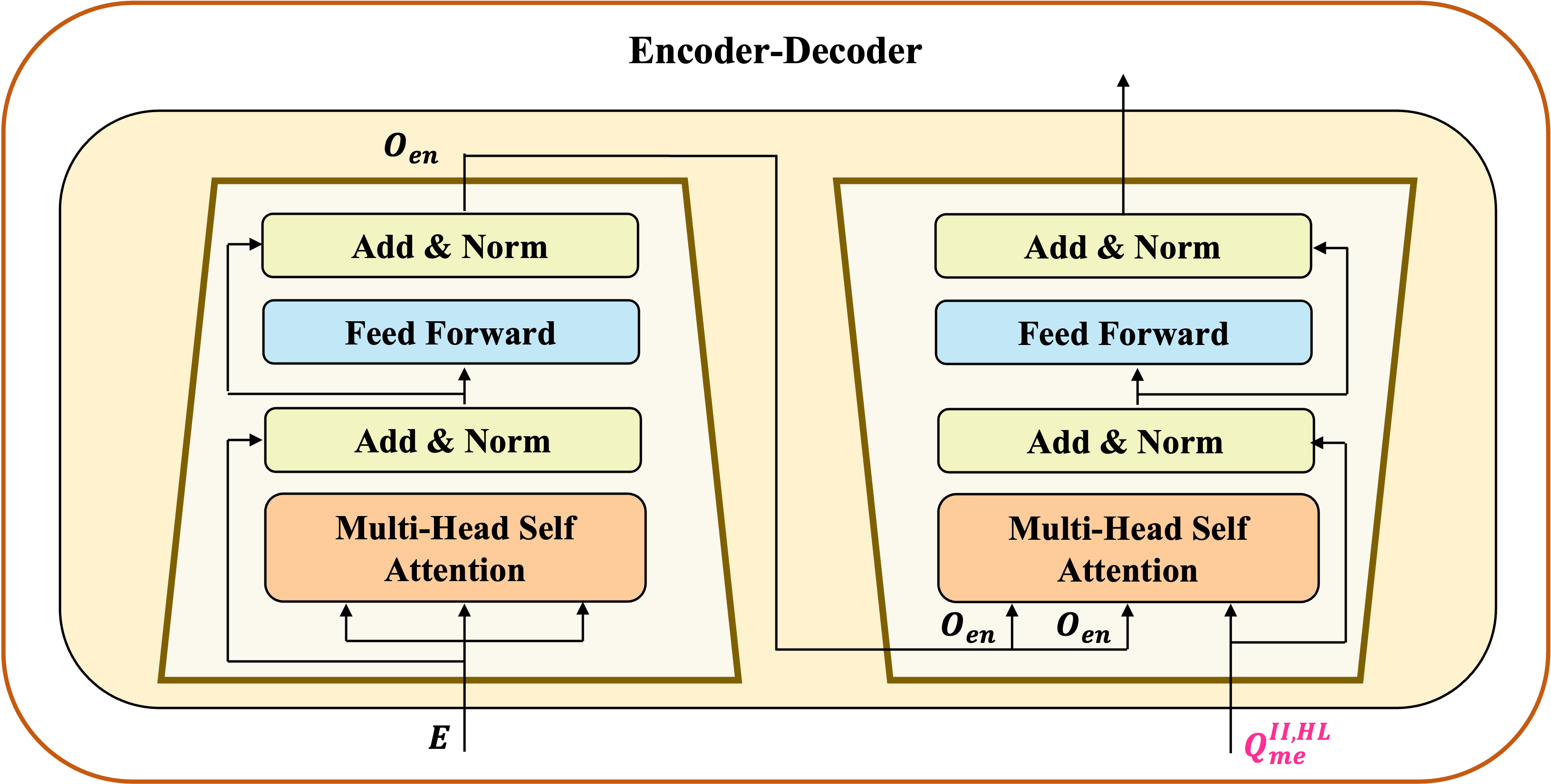}
    \caption{Diagram of Encoder-Decoder Structure in Stage \Rmnum{2}.}
    \label{fig7}
\end{figure}

\subsubsection{Encoder-Decoder Structure}
\label{enc-dec}

The encoder–decoder structure in Stage \Rmnum{2} is illustrated in Fig.\ref{fig7}. We largely follow the Transformer backbone, except that positional encoding is removed, and only the components after the cross-attention module in the decoder are retained.

\subsection{Model Optimization}
\label{Model Optimization}

Stage \Rmnum{1} of DNA is actually a multi-task learning problem. The network simultaneously outputs the predicted target $Y^{I}_{dd}$ and a set of coefficients $coefs$, which are used to construct a correction function $\mathcal{F(\cdot)}$. This function refines the original output $Y^I_{me}$ of the mechanistic model and produces a corrected result, $Y^I_{cor}=\mathcal{F}(Y^I_{me};coefs)$. Therefore, to train this model well, it is necessary to minimize the following two loss functions at the same time:
\begin{equation}
    \mathcal{L}_1 = \frac{1}{N}\sum \left \| Y^I_{dd}-Y^I_{true} \right \| ^2,
    \label{equa-18}
\end{equation}
\begin{equation}
    \mathcal{L}_2 = \frac{1}{N}\sum \left \| Y^I_{cor}-Y^I_{true} \right \| ^2,
    \label{equa-19}
\end{equation}
\noindent where, $N$ represents the total number of samples. Since different tasks vary in learning difficulty, simply summing the two loss terms may lead to training imbalance, where one task dominates the optimization process and achieves significantly better performance at the expense of the other \cite{vandenhende2021multi}. To address this issue, we follow the work of \cite{kendall2018multi} and introduce task-dependent uncertainty to adaptively weight the losses. The resulting optimization objective is formulated as follows:
\begin{equation}
    w^* = \arg \min_{w}[\frac{1}{2\sigma_1^2}\mathcal{L}_1(w)+\frac{1}{2\sigma_2^2}\mathcal{L}_2(w)+ln(\sigma_1\sigma_2)],
    \label{equa-20}
\end{equation}
\noindent here, $w$ represents the set of all parameters in the data-driven model; $\sigma_1$ and $\sigma_2$ represent the task-dependent uncertainties of the two tasks. They are learnable parameters that are updated dynamically during training. However,  when $\sigma_1$ or $\sigma_2$ becomes too small, the loss term can become negative, potentially causing training instability or failure. In this case, we adopt a more pragmatic alternative by assigning a weight ($\alpha=0.5$) to one loss term. The revised optimization objective is defined as follows:
\begin{equation}
    w^*=\arg \min_{w}[\mathcal{L}_1(w)+\alpha\mathcal{L}_2(w)].
    \label{equa-21}
\end{equation}
Stage II of DNA does not involve correction function generation, model in this phase is trained using only Eq. \ref{equa-18}.

\section{Industrial System: Three-Phase Pipeline}

\label{Modeled Industrial System}

\subsection{Three-Phase Pipeline}
In the petrochemical industries, the three-phase pipeline typically refers to a multiphase flow system capable of simultaneously transporting crude oil, natural gas, and water \cite{yaqub2020review}. Such systems serve as critical links between oil production sites, processing facilities, and transportation infrastructure, and are widely applied in scenarios such as field development, gathering and distribution, and long-distance pipeline transportation. By enabling the co-transport of oil, gas, and water without prior separation, these systems can significantly reduce the infrastructure and energy costs associated with phase separation and transportation, while enhancing production efficiency and flexibility in resource allocation \cite{zhang2006unified}. 

The mechanistic modeling of three-phase pipelines typically relies on fundamental physical conservation laws, such as the conservation of mass, momentum, and energy. These conservation equations are often coupled with empirical formulas to describe interphase interactions as well as flow-related factors such as frictional losses. During the computation process, it is essential to rigorously define the properties of each phase, such as density, viscosity, and specific heat capacity, and their dependencies on temperature and pressure. However, the composition of three-phase fluids is often complex, and the formation of emulsions (e.g., oil-in-water or water-in-oil) further complicates the system. These phenomena make it difficult to accurately characterize the relevant physical properties \cite{zhou2024modeling}. Moreover, variations in pipeline geometry and operating conditions can induce changes in flow patterns, resulting in strongly nonlinear and coupled behaviors, such as interphase interactions, energy exchange, and phase transitions. These complexities hinder the formulation and solution of core conservation equations for momentum, mass, and energy. To ensure tractability, many empirical formulas, based on specific experiments or simplified assumptions, are introduced, but their limited generalizability makes it difficult to adapt to the diverse conditions of real-world operations \cite{yaqub2020review}.

In summary, the difficulty in accurately characterizing physical properties, coupled with the structural complexity of the systems, often leads to prohibitively high modeling costs for constructing mechanistic models of three-phase pipelines, yet satisfactory performance remains elusive \cite{al2014experimental}.

\begin{table*}[]
    \caption{Description of all the datasets. For certain datasets, $Z$ denotes intermediate variables specific to each case; although some symbols may appear identical across datasets, their meanings and computational definitions are entirely distinct. }
    \centering
    \resizebox{0.97\textwidth}{!}{
    \begin{tabular}{c|c|c|c|c|c}
        \toprule 
        \multicolumn{2}{c|}{Modeled object $\mathcal{O}$} & $X$ & $Z$  & $y_{me},y_{true}$  & Unit \\ \midrule
        \multicolumn{1}{c|}{\multirow{8}{*}{\rotatebox{90}{Stage \Rmnum{1}}}}  &  Cp.g  & $T$, $Mw_g$ & $z_1,z_2,z_3,z_4,z_5$  & $Cp.g_{me}$, $Cp.g_{true}$ & $J/(kg\cdot K)$  \\ \cmidrule(lr){2-6}
        
         & Cp.o& $T,SG_o,Tb_o$ & $K,A_1,A_2,A_3,CpR$ &  $Cp.o_{me}$, $Cp.o_{true}$  & $J/(kg\cdot K)$   \\ \cmidrule(lr){2-6}
        
         &Cp.w  &$T,Mw_w$  & $z_1,z_2,z_3,z_4,z_5$  & $Cp.w_{me}$, $Cp.w_{true}$  & $J/(kg\cdot K)$ \\ \cmidrule(lr){2-6}
         
         & Rho.o & $T,SG_o$  & $alpha$  &  $Rho.o_{me}$, $Rho.o_{true}$  & $kg/m^3$  \\ \cmidrule(lr){2-6}
         
         & Rho.l & $T,P,Mw_{og},Mw_{l},x_{mol}[2],Rho.w$  & $Rho.og,z_1$  & $Rho.l_{me}$, $Rho.l_{true}$  & $kg/m^3$ \\ \cmidrule(lr){2-6}
         & Vis.w & $T$ & $z_1,z_2,z_3$  &  $Vis.w_{me}$, $Vis.w_{true}$ & $N\cdot s/mm^2$ \\ \cmidrule(lr){2-6}
         
         & Vis.o & $T,SG_o$ & $API,n$  & $Vis.o_{me}$, $Vis.o_{true}$  & $N\cdot s/mm^2$  \\ \cmidrule(lr){2-6}
         
         & Vis.l & $T,x_{mol}[2]$ & $Vis_g,Vis_w,Vis_o$  & $Vis.l_{me}$, $Vis.l_{true}$   & $N\cdot s/mm^2$  \\ \midrule \midrule
         
         \multicolumn{1}{c|}{\multirow{2}{*}{\rotatebox{90}{\Rmnum{2}}}} & HL & \rule[0.07cm]{0.8cm}{1pt} & $Q_g,Q_l,R_l,v,N_{Fr},HL_{me}$  & $HL_{me},HL_{true}$  & \% \\ \cmidrule(lr){2-6}
         & P & \rule[0.07cm]{0.8cm}{1pt} & $Rho_g,Rho_l,Mw_l,phi,v_{sg},z_1,z_2,v,Rho,z_3,M_{flow},P_{me}$  &  $P_{me},P_{true}$ & $kPa$ \\

        \bottomrule
        
    \end{tabular}
    }
    \label{table-4}
\end{table*}

\subsection{Datasets}

As described in the main text (Section \ref{task definition}), we constructed a mechanistic model of the three-phase pipeline based on the Beggs–Brill theory \cite{beggs1973study,chaves2022comparative} and ran it under the same operating conditions $X$ as those of the real-world three-phase pipeline from which our data were collected, in order to obtain the mechanistic predictions ($Y^I_{me},Y^{II}_{me}$) together with a set of key intermediate variables ($Z^I$). Using these data, we formed eight datasets for Stage \Rmnum{1} and two datasets for Stage \Rmnum{2}.

In the three-phase pipeline system, we select eight physical properties as Stage \Rmnum{1} modeled objects: gas-phase specific heat capacity \textbf{Cp.g}, oil-phase specific heat capacity \textbf{Cp.o}, water-phase specific heat capacity \textbf{Cp.w}, oil-phase density \textbf{Rho.o}, liquid-phase density \textbf{Rho.l}, water-phase viscosity \textbf{Vis.w}, oil-phase viscosity \textbf{Vis.o}, and liquid-phase viscosity \textbf{Vis.l}. In Stage \Rmnum{2}, we further model two downstream objects—liquid holdup \textbf{HL} and pressure \textbf{P} at the pipeline outlet—\textbf{which are derived from the aforementioned physical properties through additional calculations}. As a result, this study involves a total of ten modeled objects, each associated with a distinct dataset. \textbf{It is important to note that although these datasets originate from the same industrial system, they differ significantly in terms of input-output variables and data distributions.} Detailed information about these datasets is provided in Tab.\ref{table-4}.

\subsection{Mechanistic Model}
\subsubsection{Introduction}
In three-phase pipeline, the dominant flowing media consist of \textbf{gas}, \textbf{water}, and \textbf{oil}. The oil and water phases are immiscible, and no mass transfer occurs between the gas and water phases. The dissolution and separation processes between the oil and gas phases are assumed to be instantaneous. The definitions of gas, water, and oil used in this work are given below:

\begin{itemize}
    \item The gas phase is composed of methane $CH_4$.
    \item The water phase consists of water $H_2O$.
    \item The oil phase is represented by a pseudo-component of representative oil (with a specific gravity $SG_o=0.84$ and a boiling point $Tb_o = 569.65K$) mixed with a small amount of dissolved methane.
    \item The liquid phase comprises both the oil and water phases.
\end{itemize}

Consequently, during the Stage \Rmnum{1} computation, the gas-phase physical property ($Cp.g$) is taken as that of methane; the water-phase physical properties ($Cp.w, Vis.w$) correspond to those of water; the oil-phase physical properties ($Cp.o, {Rho}.o, {Vis}.o$) are calculated as the mixture properties of the representative oil pseudo-component and dissolved methane; and the liquid-phase physical properties (${Rho}.l,Vis.l$) are defined as the properties of the liquid-phase mixture.

\subsubsection{Mechanistic Model of \textbf{Cp.g}}

The mechanistic model of Cp.g can be expressed as follows:

\begin{equation}
    Cp.g = [VapCp[1]+VapCp[2]\times(\frac{VapCp[3]}{T\times \sin\frac{VapCp[3]}{T}})^2+VapCp[4]\times (\frac{VapCp[5]}{T\times \cos\frac{VapCp[5]}{T}})^2]/Mw_g
    \label{a-equa7}
\end{equation}

\noindent where $T$ denotes the actual temperature, $\{VapCp[i]\}_{i=1}^5$ represents the specific heat parameter of the gas phase, and $Mw_g$ is the relative molecular weight of the gas phase (methane). The intermediate variables in this dataset are defined as follows:

\begin{equation}
    z_1=T\times \sin \frac{VapCp[3]}{T},
\end{equation}
\begin{equation}
    z_2 = T \times \cos \frac{VapCp[5]}{T},
\end{equation}
\begin{equation}
    z_3 = VapCp[2]\times(\frac{VapCp[3]}{z_1})^2,
\end{equation}
\begin{equation}
    z_4 = VapCp[4]\times (\frac{VapCp[5]}{z_2})^2,
\end{equation}
\begin{equation}
    z_5 = VapCp[1]+z_3+z_4.
\end{equation}

\subsubsection{Mechanistic Model of \textbf{Cp.o}}

The mechanistic model of Cp.o can be expressed as follows:

\begin{equation}
    Cp.o=4.1868\times (A_1+A_2\times TR+A_3 \times TR^2),
\end{equation}

\noindent where $A_1,A_2,A_3,TR$ are fitting parameters used in engineering practice with no explicit physical meaning. The intermediate variables in this dataset are defined as follows:

\begin{equation}
    A_1 = -1.17126+(0.023722+0.024907\times SG_o)\times K+\frac{1.14982-0.046535\times K}{SG_o},
\end{equation}
\begin{equation}
    A_2 =10^{-4}\times(1+0.82463\times K)\times(1.12172-\frac{0.27634}{SG_o}),
\end{equation}
\begin{equation}
    A_3= -10^{-8}\times(1+0.82463\times K)\times(2.9027-\frac{0.70958}{SG_o}),
\end{equation}
\begin{equation}
    K=\frac{(1.8\times Tb_{o})^{1/3}}{SG_o},
\end{equation}
\begin{equation}
    TR=1.8\times T,
\end{equation}
\begin{equation}
    CpR=A_1+A_2\times TR+A_3\times TR^2,
\end{equation}

\noindent where $K, SG_o, Tb_o$ denote the characterization factor, specific gravity, and boiling point of the pseudo-component of representative oil, respectively. The correspondence between these variables and those in Fig.\ref{fig1} of the main paper is as follows: $\{x_1,x_2,x_3,z_1,z_2,z_3,z_4,z_5\}$ in Fig.~1 correspond to $\{{SG},Tb_o,T,K,A_1,A_2,A_3\}$, respectively.

\subsubsection{Mechanistic Model of \textbf{Cp.w}}

The mechanistic model of Cp.w can be expressed as follows:

\begin{equation}
    Cp.w=[LiqCp[1] +LiqCp[2]\times T+LiqCp[3] \times T^2+LiqCp[4] \times T^3+LiqCp[5] \times T^4]/Mw_w,
\end{equation}

\noindent where $\{LiqCp[i]\}_{i=1}^5$ represents the specific heat parameter of the water phase. The intermediate variables in this dataset are defined as follows:

\begin{equation}
    z_1 = LiqCp[2]\times T,
\end{equation}
\begin{equation}
    z_2=LiqCp[3] \times T^2,
\end{equation}
\begin{equation}
    z_3 = LiqCp[4] \times T^3,
\end{equation}
\begin{equation}
    z_4= LiqCp[5] \times T^4,
\end{equation}
\begin{equation}
    z_5 = LiqCp[1]+z_1+z_2+z_3+z_4.
\end{equation}

\subsubsection{Mechanistic Model of \textbf{Rho.o}}

The mechanistic model of Rho.o can be expressed as follows:

\begin{equation}
    Rho.o=\frac{SG_o\times1000}{1+alpha\times (T-20)},
\end{equation}
\begin{equation}
    alpha=(3.083-2.638 \times SG_0) \times 10^{-3}
\end{equation}

\subsubsection{Mechanistic Model of \textbf{Rho.l}}

The mechanistic model of Rho.l can be expressed as follows:

\begin{equation}
    Rho.l=\frac{1}{\frac{x_{mol}[2]}{Rho.w}+\frac{1-x_{mol}[2]}{Mw_{og}\times Rho.og}}\times Mw_l,
\end{equation}

\noindent where $Mw_{og}, Mw_{l}, x_{mol}[2],{Rho}.w, Rho.og$ denote the relative molecular weight of the oil–gas mixture, the relative molecular weight of the liquid phase, the molar fraction of the water phase in the liquid phase, the density of the water phase, and the density of the oil-gas mixture respectively. The computation of ${Rho}.{og}$ and another intermediate variable $z_1$ is given as follows:

\begin{equation}
    Rho.{og}=\frac{Rho.o+Rs\times(Rs\times(0.00379\times do-0.00393)-4.08779\times do+4.43818)×1.29}{0.972+0.000147\times(5.62\times Rs\times(\frac{dg}{do})^{0.5}+2.25\times T+40)^{1.175}},
\end{equation}
\begin{equation}
    Rs = 2.2\times dg\times (\frac{P}{10}\times 10^{(\frac{1.77}{do}-0.001638\times T-1.67)})^{1.205},
\end{equation}
\begin{equation}
    do=\frac{Rho.o}{Rho.w}(T=288.15K),
\end{equation}
\begin{equation}
    dg=0.86,
\end{equation}
\begin{equation}
    z_1 = \frac{1}{\frac{x_{mol}[2]}{Rho.w}+\frac{1-x_{mol}[2]}{Mw_{og}\times Rho.og}},
\end{equation}

\noindent where $T, P, R_s, d_g,d_o$ denote the actual temperature, actual pressure, the gas–oil ratio of oil (unit: $\frac{m^3}{m^3}$), the relative density of methane with respect to air, and the oil–water density ratio at a temperature of 288.15 K, respectively.

\subsubsection{Mechanistic Model of \textbf{Vis.w}}

The mechanistic model of Vis.w can be expressed as follows:

\begin{equation}
    Vis.w=\exp{(LiqVis[1]+\frac{LiqVis[2]}{T}+LiqVis[3]\times ln{(T)}+LiqVis[4]\times T^{LiqVis[5]})},
\end{equation}

\noindent where $\{LiqVis[i]\}_{i=1}^5$ represents liquid-phase viscosity parameter. The intermediate variables in this dataset are defined as follows:

\begin{equation}
    z_1 = \frac{LiqVis[2]}{T},
\end{equation}
\begin{equation}
    z_2 =LiqVis[3]\times ln{(T)},
\end{equation}
\begin{equation}
    z_3 = LiqVis[4]\times T^{LiqVis[5]}.
\end{equation}

\subsubsection{Mechanistic Model of \textbf{Vis.o}}

The mechanistic model of Vis.o can be expressed as follows:

\begin{equation}
    Vis.o=3.141\times10^7 \times |1.8\times T-460|^{-3.444} \times (log_{10}(API))^n,
\end{equation}

\noindent $API$ is another representation of the specific gravity of the pseudo-component of representative oil, and $n$ is a fitting parameter

\begin{equation}
    n =10.313 \times log_{10}(|1.8\times T-460|)-36.447,
\end{equation}
\begin{equation}
    API = \frac{141.5}{SG_o}-131.5,
\end{equation}

\subsubsection{Mechanistic Model of \textbf{Vis.l}}

The mechanistic model of Vis.l can be expressed as follows:

\begin{equation}
    Vis.l=x_{mol}[1]\times Vis.g+x_{mol}[2]\times Vis.w+x_{mol}[3]\times Vis.o
\end{equation}

\noindent where $x_{mol}[1], x_{mol}[2], x_{mol}[3]$ denote the molar fractions of the dissolved gas, water phase, and oil phase in the liquid phase, respectively.

\begin{figure}[]
\centering
\includegraphics[width=\linewidth]{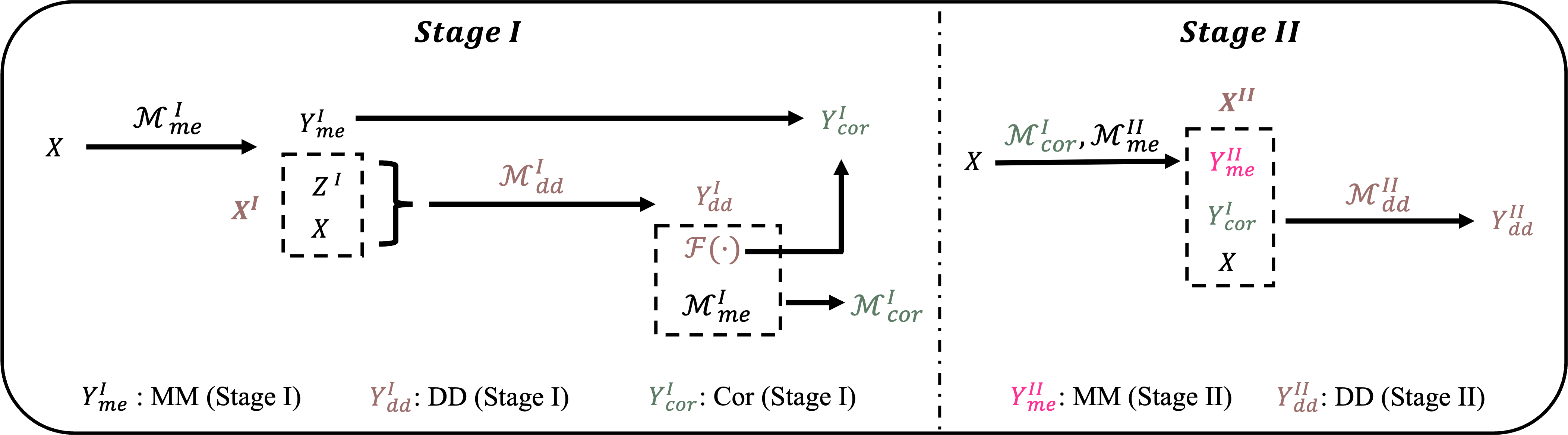}
\caption{Data flow in this work.}
\label{fig8}
\end{figure}

\subsection{The Data Flow in Three-Phase Pipeline}
\label{dataflow}

In this work, the modeling targets in Stage \Rmnum{2} include the liquid holdup (HL) in the three-phase pipeline and the pressure (P) at the pipeline outlet. Mechanistically, these variables are computed based on the physical properties obtained in Stage \Rmnum{1} through further calculations. To present the logic of this work more clearly and intuitively, the complete data flow is illustrated in Fig.\ref{fig8}.

In Stage \Rmnum{1}, the operational conditions of the three-phase pipeline collected from the real world (e.g., $T_{inlet}$, $P_{inlet}$, $HL_{inlet}$, etc.) serve as the raw inputs $X$ to our model. We then run the mechanistic model $\mathcal{M}_{me}^{I}$ under exactly the same operating conditions to obtain the mechanistic outputs $Y_{me}^{ I}$ and $Z^{I}$. The set $X^I:\{X, Z^{I}\}$ is subsequently used as input to the Stage \Rmnum{1} data-driven model $\mathcal{M}_{dd}^{I}$, producing the data-driven outputs $Y_{dd}^{I}$ and a set of correction formulas $\mathcal{F}(\cdot)$. Applying $\mathcal{F}(\cdot)$ to the original mechanistic model $\mathcal{M}_{me}^{I}$ yields the Stage \Rmnum{1} corrected mechanistic model $\mathcal{M}_{cor}^{I}$. $Y_{{me}}^I$ is corrected by the correction functions to obtain $Y_{{cor}}^I$, which is more accurate than $Y_{me}^{I}$.

In Stage II, we run $\mathcal{M}_{cor}^{I},\mathcal{M}_{me}^{II}$ under the same real-world operating conditions $X$, obtaining
the outputs $Y^{II}_{me}$. Then, the set $X^{II}:\{Y_{me}^{II},Y^I_{cor},X\}$ are used as inputs to the Stage \Rmnum{2} data-driven model $\mathcal{M}_{dd}^{II}$ to obtain the predictions $Y_{dd}^{II}$.

This completes the full data flow in this work. In this process, $Y_{me}^{I}$, $Y_{dd}^{I}$, and $Y_{cor}^{I}$ correspond to the Stage \Rmnum{1} MM, DD, and Cor in Tab. \ref{table1} of the main paper; while $Y_{me}^{II}$ and $Y_{dd}^{II}$ correspond to the Stage \Rmnum{2} MM and DD in the same table.

\section{Industrial System: Acentric Factor}

\label{Modeled Industrial System: Acentric Facotr}

\subsection{Acentric Factor and the Mechanistic Model}
The acentric factor ($\omega$) is an unmeasurable dimensionless parameter used to describe the extent to which real gas deviate from ideal gas behavior. It plays a significant role in thermodynamics and fluid mechanics \cite{biswas2023predicting}. Although there are numerous methods for calculating the acentric factor, such as the Edmister method \cite{edmister1984applied}, Nath method \cite{nath1976choice}, and Lee-Kesler method \cite{lee1975generalized}, these mechanistic formulas are derived from empirical data and thus have limitations in their applicability. When the values fall within the range for which these formulas are valid, the deviations are minimal; however, outside this range, the deviations can be significant \cite{abakporo2021general}. Among all these methods, the Lee-Kesler method is the most commonly used due to its simple structure. Therefore, we adopt the Lee-Kesler method as the mechanistic model for the acentric factor. The Lee-Kesler method is presented as follows:
\begin{equation}
    \psi =  - \ln_{}{(\frac{{P}_{c}}{1.01325}}-5.92714+\frac{6.09648}{{T}_{br}}  + 1.28862\times \ln_{}{{T}_{br}}-0.169347\times {T}_{br}^{6},
\end{equation}
\begin{equation}
    \phi =  15.2518-\frac{15.6875}{{T}_{br}}-13.4721\times \ln_{}{{T}_{br}} + 0.43577\times {T}_{br}^{6},
\end{equation}
\begin{equation}
    \rho =  -7.904+0.1352\times {K}_{w}-0.007465\times {K}_{w}^{2}+8.359\times {T}_{br}+ 
                \frac{1.408-0.01063\times{K}_{w}}{{T}_{br}},
\end{equation}
\begin{equation}
    \displaystyle
    \omega=\begin{cases}
        \frac{\psi}
        {\phi},&{T}_{br}\leq 0.8, \\
        \\
        \rho, &{T}_{br}>0.8,\end{cases}   
\end{equation}%

\noindent where $P_c$, $T_{br}$, and $K_w$ represent the critical pressure, reduced boiling temperature, and characteristic factor, respectively. All of them can be calculated from the boiling point (${T}_b$) and specific gravity ($SG$), with the specific calculation formulas to be introduced later.

\subsection{Dataset}
During dataset acquisition, a pseudo-component approach \cite{riazi2005characterization} is applied to petroleum with boiling points ranging from 300K to 1100K, resulting in pseudo-components characterized by their boiling point (${T}_b$) and specific gravity ($SG$). Each pseudo-component represents a sample in this dataset, with the ground truth $\omega_{gt}$ obtained using Aspen Plus, a famous and efficient simulation software \cite{liu2022process,al2022aspen}. The input for each sample consists of six variables \{${T}_b$, $SG$, $K_w$, ${T}_{br}$, $T_c$, $P_c$\}, where the characteristic factors ($K_w$), reduced boiling point (${T}_{br}$), critical temperature ($T_c$), and critical pressure ($P_c$) for each pseudo-component can be calculated from $T_b$ and $SG$ using the equations provided below \cite{riazi2005characterization}:
\begin{equation}
{T}_{br}=\frac{{T}_{b}}{{T}_{c}},    
\end{equation}
\begin{equation}
    {K}_{w}=\frac{{(1.8{T}_{b})}^{1/3}}{{SG}},
\end{equation}
\begin{equation}
    T_c = 19.06232\times T_b^{0.58848} \times SG^{0.3596},
\end{equation}
\begin{equation}
    P_c = 5.53027*10^7*{T_b}^{-2.3125}*SG^{2.3201} .
\end{equation}

\begin{figure}
    \centering
    \includegraphics[width=0.6\linewidth]{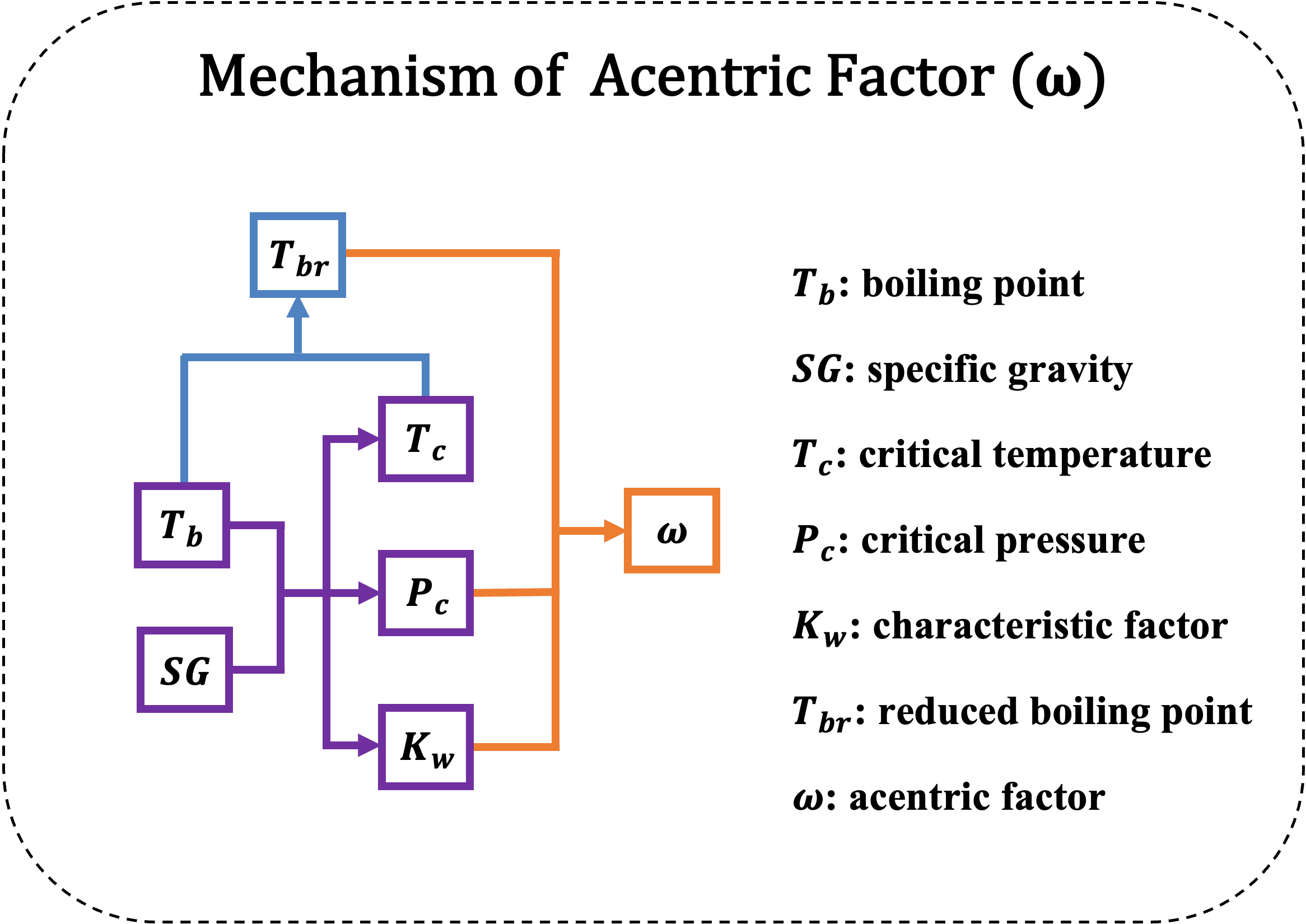}
    \caption{The mechanism of acentric factor.}
    \label{fig15}
\end{figure}

The underlying mechanism of the acentric factor is illustrated in Fig.\ref{fig15}. The mechanistic model of the acentric factor is composed of explicit functions and exhibits limited accuracy; therefore, in this experiment, it is modeled using the Stage I approach of DNA, which also improves the performance of the mechanistic model itself.

\section{Experiments}
\label{exper}

\subsection{Baselines}
\label{baselines}
As described in Section\ref{task definition}, the industrial digital twin task considered in this work can be formulated as follows:

\begin{equation}
    \min_{M} \; \mathbb{E}_{x \sim \mathcal{X}} \left[ \left| \mathcal{M}(x) - \mathcal{O}(x) \right| \right],
\end{equation}

\noindent where $\mathbb{E}[\cdot]$ denotes the mathematical expectation and $\mathcal{X}$ denotes the distribution of operating conditions. Therefore, the task considered in this work is essentially a regression problem. To thoroughly evaluate DNA, we select four categories of models with strong regression capabilities: tree-based, transformer-based, graph-based, and other representative models. Below, we introduce each model individually:

\begin{itemize}
    \item LightGBM \cite{ke2017lightgbm} is a boosting decision tree, with gradient-based one-side sampling and exclusive feature bundling.
    \item Transformer \cite{vaswani2017attention} is a sequence-based model, and here, we utilize only its encoder component for industrial modeling.
    \item FlowformerMH \cite{wu2022flowformerlinearizingtransformersconservation} is a linear transformer that models global interactions through flow conservation and efficient attention mechanisms, reducing the computational complexity of conventional self-attention.
    \item iTransformer \cite{liu2024itransformerinvertedtransformerseffective} inverts the conventional Transformer architecture by treating individual variables as tokens, enabling the model to capture multivariate correlations more effectively.
    \item MultiPatchFormer \cite{naghashi2025multiscale} is a multi-scale patch-based Transformer that captures underlying patterns at different resolutions by modeling patches with multiple scales.
    \item GATv2 \cite{velivckovic2017graph} leverages attention mechanisms to dynamically assign different importance weights to a node’s neighbors ,enabling more effective learning on graph-structured data.
    \item DGDL \cite{zhu2022dynamic} uses a dynamic graph to realize adaptive learning and automatic inference and aggregate node feature representations by a multihop attention graph convolution network.
    \item TGCNS \cite{kong2023collaborative} emphasis lies in using a graph structure learning module to learn potential intervariable relationships from data.
    \item RaDA \cite{chen2024residual} is a residual-aware deep attention graph convolutional network which uses a residual-aware connection module to reduce data uncertainty and alleviate over-smoothing.
    \item TimesNet \cite{wu2022timesnet} transforms one-dimensional input variations into two-dimensional tensors and employs convolutional operations to capture complex patterns and interactions, providing strong representation and regression capabilities.
    \item TimeXer \cite{wang2024timexer} is a Transformer-based architecture that employs patch-wise modeling and cross-attention to capture complex dependencies and interactions among variables, enabling effective representation learning for regression tasks.
    \item Umixer \cite{ma2024umixerunetmixerarchitecturestationarity} is a U-Net style MLP-based architecture that employs hierarchical mixing operations to capture multiscale patterns and learn effective representations for regression tasks.
\end{itemize}

\begin{table}[]
    \caption{General hyperparameters setting.}

    \centering
    \begin{tabular}{c|c|c|c|c|c|c}
            \toprule
         Stage &  Dataset & Batch Size& Dim & lr& Max Training Epochs& Early Stopping at Epoch \\\midrule
         \multirow{8}{*}{Stage \Rmnum{1}}& Cp.g & 256 &32& 1e-4 & 2000 &1733\\
                        & Cp.o & 256 &32& 1e-4 & 2000 &1140\\
         & Cp.w & 256 &32& 5e-4 & 2000 &502\\
         & Rho.o & 256 &32& 1e-4 & 2000 &1236\\
         & Rho.l & 256 &32& 5e-4 & 2000 &2000\\
         & Vis.w & 256 &32& 1e-4 & 2000 &1636\\
         & Vis.o & 256 &32& 5e-4 & 2000 &1598\\
         & Vis.l & 256 &32& 1e-4 & 2000 &1238\\ 
         & $\omega$ & 256 &32& 1e-3 & 2000 &298\\ \midrule
         \multirow{2}{*}{Stage \Rmnum{2}}& HL & 256 &32 & 1e-3 & 2000 &439\\ 
         & P & 256 &64  & 1e-3 & 2000 &940\\ \bottomrule
    \end{tabular}
    \label{table5}
\end{table}

\subsection{Implementation Details}
\label{d1}

We split all datasets into training, validation, and testing sets with a ratio of 7:2:1. To prevent overfitting, all models are trained using an early stopping strategy. The maximum number of training epochs is uniformly set to 2000, while the actual stopping epoch varies across different datasets. All experiments are implemented in PyTorch \cite{paszke2019pytorch} and conducted on a machine equipped with four NVIDIA GeForce RTX 4090 GPUs. We adopt the Adam optimizer \cite{kingma2014adam} for all experiments. Detailed hyperparameter settings for each dataset are reported in Tab. \ref{table5}.

\subsection{Choice of Correction Function Form}
\label{correctino function}
As described in the main paper, the predefined function form for each dataset and the correction functions ultimately obtained by our model are summarized in Tab.\ref{table6}.

\renewcommand{\arraystretch}{1.5}
\begin{table*}[]
    \caption{Correction functions of all the dataset in Stage \Rmnum{1}}
    \centering
    \resizebox{\textwidth}{!}{
    \begin{tabular}{c|c|c|c|c}
        \toprule
        Dataset & $n$ & $ln$ & $gate$ & Correction Function \\ \midrule
         Cp.g& 2    & \ding{55} &\ding{51} &$y_{cor}=0.8342991\times1.472976 +0.52315223\times0.8895244\times y_{me}+0.30744648\times0.0457618\times y_{me}^2$ \\ \midrule
         Cp.o& 2    & \ding{55} & \ding{51}  & $y_{cor}=0.7068145\times1.2715807+0.5873085\times0.5926248\times y_{me}+0.59426093\times0.17929147\times y_{me}^2$ \\ \midrule

         Cp.w& 0    & \ding{51} &\ding{55} &$y_{cor}=0.82825834+1.9789125\times ln(y_{me}+1.278508)$  \\ \midrule
         Rho.o & 2 & \ding{51} & \ding{51} &
                            $\begin{aligned}
                            y_{{cor}} =\;& 0.4679038 \times 0.42148522 + 0.6566006 \times 0.4409263 \times y_{{me}} 
                            + 0.6441115 \times 0.55001056 \times y_{{me}}^{2} \\
                            &+ 0.4212044 \times
                            \ln(|-0.82146084\times y_{{me}} - 0.7055963|)
                            \end{aligned}$ \\\midrule
         Rho.l& 2    & \ding{51} & \ding{55}&$y_{cor}=0.05121769+0.87679595\times y_{me}-0.5762174\times y_{me}^2+ ln(|1.5252085\times y_{me}+0.34641176|)$  \\\midrule
         Vis.w& 2    & \ding{51} & \ding{55}&$y_{cor}=0.4235715+0.5907204\times y_{me}-0.26162362\times y_{me}^2+ln(|0.3140283\times y_{me}+0.6718855|)$   \\\midrule
         Vis.o& 3    & \ding{51} & \ding{51}&
         
         $\begin{aligned}
                            y_{cor} =\;& 0.6880405\times2.018896+0.49424344\times1.8307866\times y_{me}-0.43409765\times1.4872801\times y_{me}^2 \\
                            &+0.19532314\times 0.73345625\times y_{me}^3+0.14005646\times ln(|-2.045505\times y_{me}+3.0183353|)
                            \end{aligned}$
          \\\midrule
         Vis.l& 3    & \ding{51} & \ding{55}& $y_{cor}=2.210485-3.1021433\times y_{me}+1.5562956\times y_{me}^2-0.24822733\times y_{me}^3+ln(|-4.001664\times y_{me}+0.9682727|)$ \\\midrule
         $\omega$& 2    & \ding{55} & \ding{55}& $y_{cor}=0.17715707+0.4112167\times y_{me}+0.43530726\times y_{me}^2$ \\
         \bottomrule
    \end{tabular}
    }

    \label{table6}
\end{table*}

\subsection{More Discussion on Stage I Experimental Results of the Three-Phase Pipeline}
\label{more analysis on Stage I}

An important point that warrants clarification is that, in Stage I, it is not necessary to directly compare the results of corrected mechanistic model (Cor) with those of the data-driven model (DD). In industrial practice, data-driven models are often difficult to deploy at scale due to their black-box nature, particularly in safety-critical applications, whereas mechanistic models remain widely adopted because they are grounded in physical principles and offer physical transparency and stability. However, mechanistic models often suffer from limited accuracy due to their complex formulation, reliance on expert knowledge, and idealizing assumptions. This perspective motivates us to develop the DNA architecture, where the enhanced mechanistic model is intended for industrial deployment, filling the gap left by data-driven models that are difficult to deploy at scale while partially overcoming the accuracy limitations of itself. Therefore, a substantial improvement of the enhanced mechanistic model over the original mechanistic model indicates that our objective has been achieved, regardless of whether it outperforms the data-driven model. In this context, comparing $\mathcal{M}^I_{{cor}}$ with $\mathcal{M}^I_{{dd}}$ is not essential: $\mathcal{M}^I_{{cor}}$ does not need to outperform $\mathcal{M}^I_{{dd}}$; improvement over $\mathcal{M}^I_{{me}}$ is sufficient.

\begin{table}[]
    \caption{Comparison of model parameters, FLOPs, and batch inference time on the Cp.w dataset.}
    \centering
    \begin{tabular}{c|c|c|c}
    \toprule
         Models &  Parameters&  FLOPs&  Inference Time (ms) \\\midrule
         DNA & \textbf{8726} &\textbf{67540}  &   32.85    \\
         Transformer& 65476 &213984 &  12.38     \\
         FlowFormer&  16253 &243241 &  12.95  \\
         iTransformer& 135460 &  89264 &  19.28       \\
         MPFormer& 77444 & 210288 &   81.78    \\
         GATV2&   32722  &  134752 &    14.70     \\
         RaDa&  13652& 67730 &  \textbf{12.11}    \\
         TimesNet&  44264   &83008 &   24.27       \\
         TimeXer&    46628  & 114784    & 20.40  \\
         Umixer& 105758 & 89768 &   21.67       \\
         \bottomrule
    \end{tabular}

    \label{table7}
\end{table}

\subsection{Further Analysis of Model Complexity and Efficiency}
\label{more results}

We report model complexity and efficiency metrics in Tab.\ref{table7}. In terms of total number of parameters and FLOPs required for a single forward inference, DNA shows a clear advantage over other baselines, indicating that our model is highly lightweight. Moreover, due to its reduced FLOPs, DNA theoretically possesses the potential for efficient inference.

However, as shown by the inference time in Tab.\ref{table7}, DNA does not yet achieve the ideal inference efficiency. This is because, in MASA, we assign different attention scopes to different variables based on the underlying mechanism of the modeled object. Each variable performs computations within its own attention scope to produce the corresponding updated representations. Since the attention scopes differ across variables, it is difficult to perform fully parallel computation as in vanilla self-attention, which can generate updated representations for all variables simultaneously. Instead, each variable’s representation must be computed sequentially according to its assigned scope, which significantly increases the inference time. This explains the current suboptimal inference efficiency of DNA.

Nonetheless, precisely because different attention scopes are assigned to different variables, the data-driven model constructed by DNA both adheres to the intrinsic mechanism of the modeled object and avoids unnecessary attention computations. As a result, when replacing MHSA with MASA, the FLOPs are substantially reduced, as shown in Tab.\ref{table8}.

In summary, the current suboptimal inference efficiency of DNA is primarily due to the fact that we have not yet found an effective way to parallelize MASA. Addressing this implementation-related issue will be one of our key focuses in future work.

\begin{table}[]
    \caption{FLOPs comparison between DNA (with MASA) and DNA (with MHSA) on Stage \Rmnum{1} datasets.}
    \centering
    \begin{tabular}{c|c|c|c}
    \toprule
         Dataset& FLOPs(DNA with MASA) & FLOPs(DNA with MHSA) & FLOPs Reduction (\%)\\ \midrule
         Cp.g& 67540&79892&15.46\\
         Cp.o&70740&79892&11.46\\
         Cp.w&67412&79892&15.62\\
         Rho.o&45028&51620&12.77\\
         Rho.l&72560&87280&16.87\\
         Vis.w&53760&58496&8.10\\
         Vis.o&53632&58496&8.32\\
         Vis.l&59164&65500&9.67\\
         \bottomrule
    \end{tabular}

    \label{table8}
\end{table}

\subsection{Visualization}
To intuitively demonstrate the improvement brought by the Stage \Rmnum{1} correction, we visualize the results at 16 representative positions from a randomly selected case. Due to space constraints, the main paper only presents visualizations for four datasets (Cp.g, Vis.l, HL, P). Here, we supplement the remaining six visualization results from the three-phase pipeline experiments, as well as those from the acentric factor experiments, as shown in the figures below.

\begin{minipage}[t]{0.3\textwidth}
    \centering
    \includegraphics[width=\textwidth]{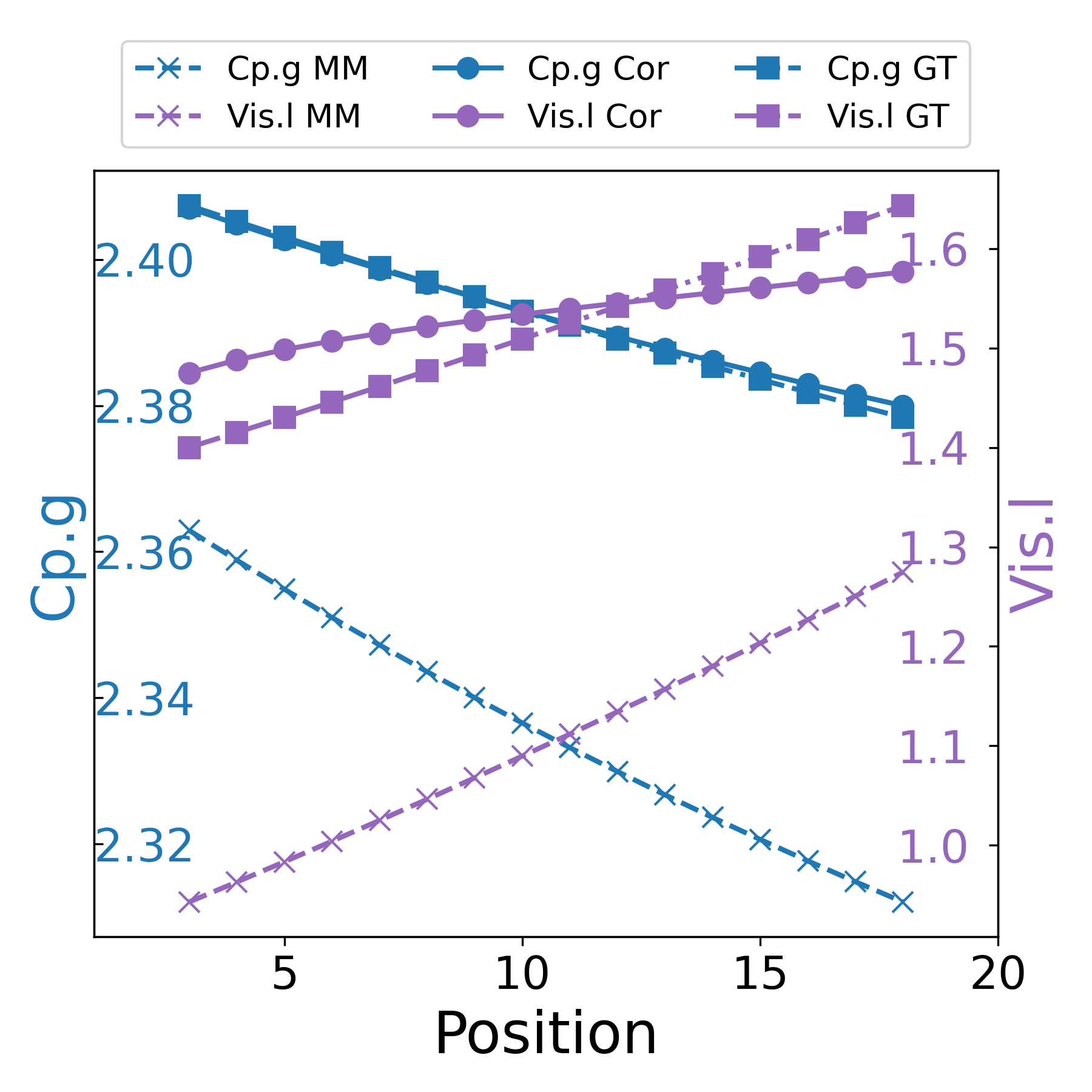}
    \label{fig9}
\end{minipage}
\hfill
\begin{minipage}[t]{0.3\textwidth}
    \centering
    \includegraphics[width=\textwidth]{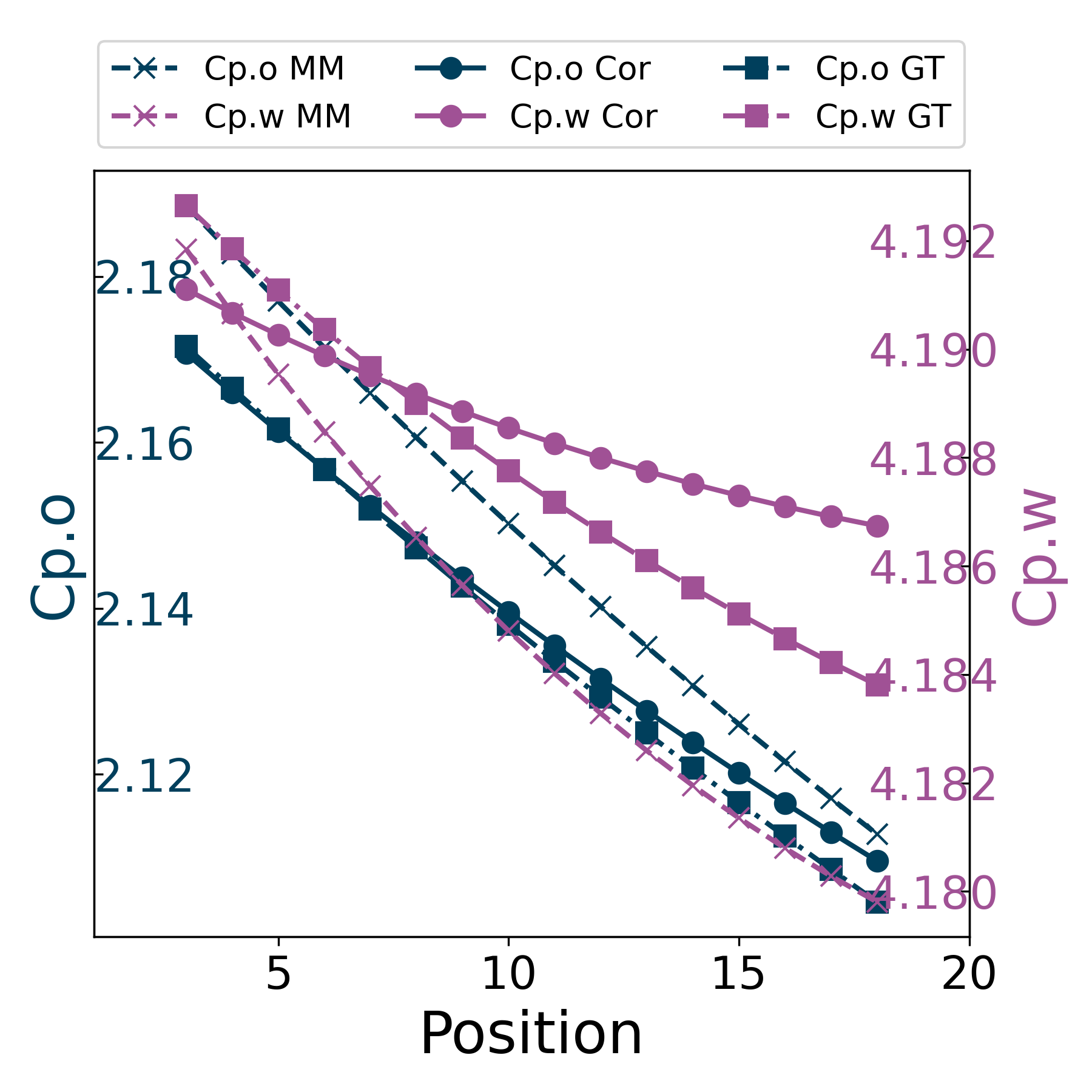}
    \label{fig10}
\end{minipage}
\hfill
\begin{minipage}[t]{0.3\textwidth}
    \centering
    \includegraphics[width=\textwidth]{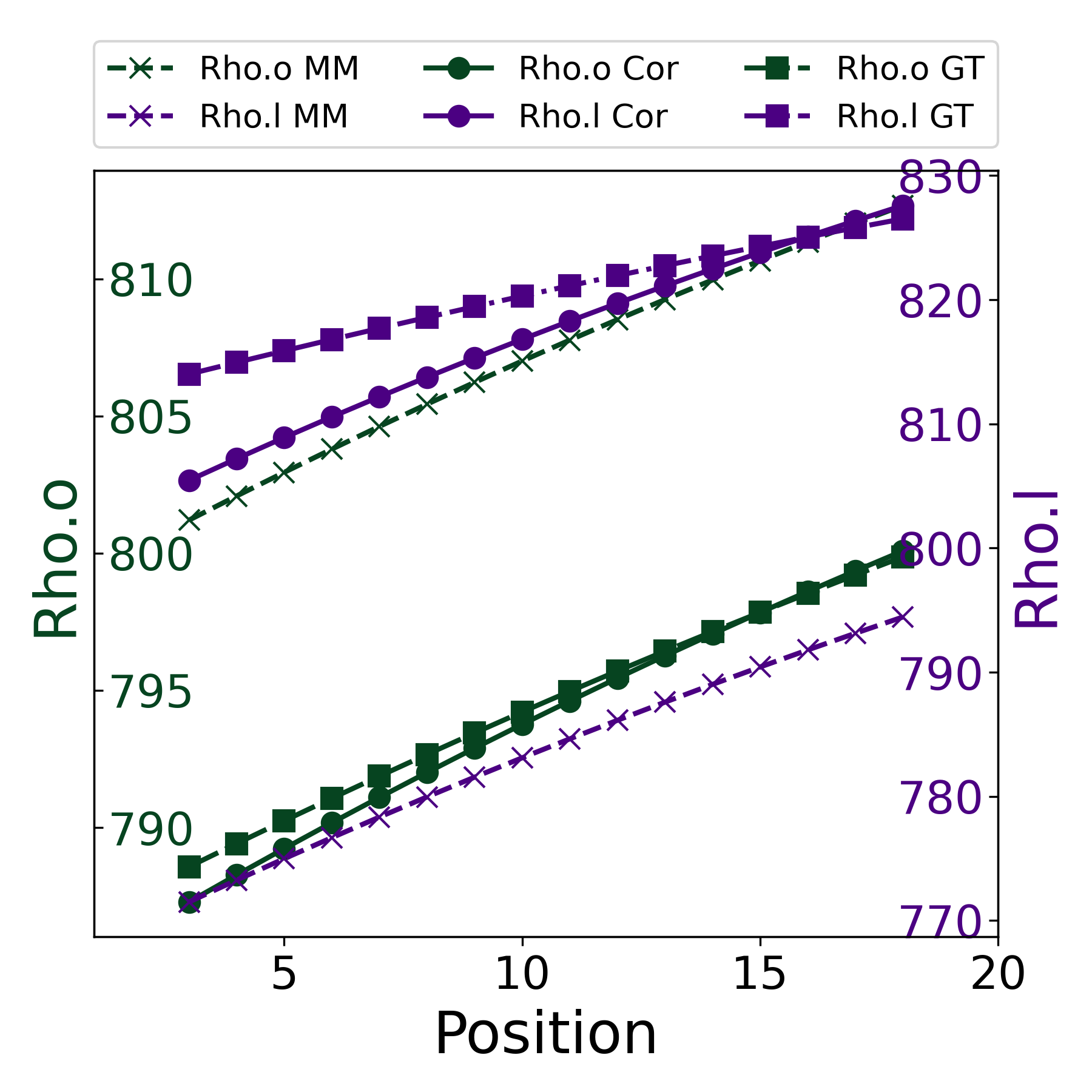}
    \label{fig11}
\end{minipage}

\begin{minipage}[t]{0.3\textwidth}
    \centering
    \includegraphics[width=\textwidth]{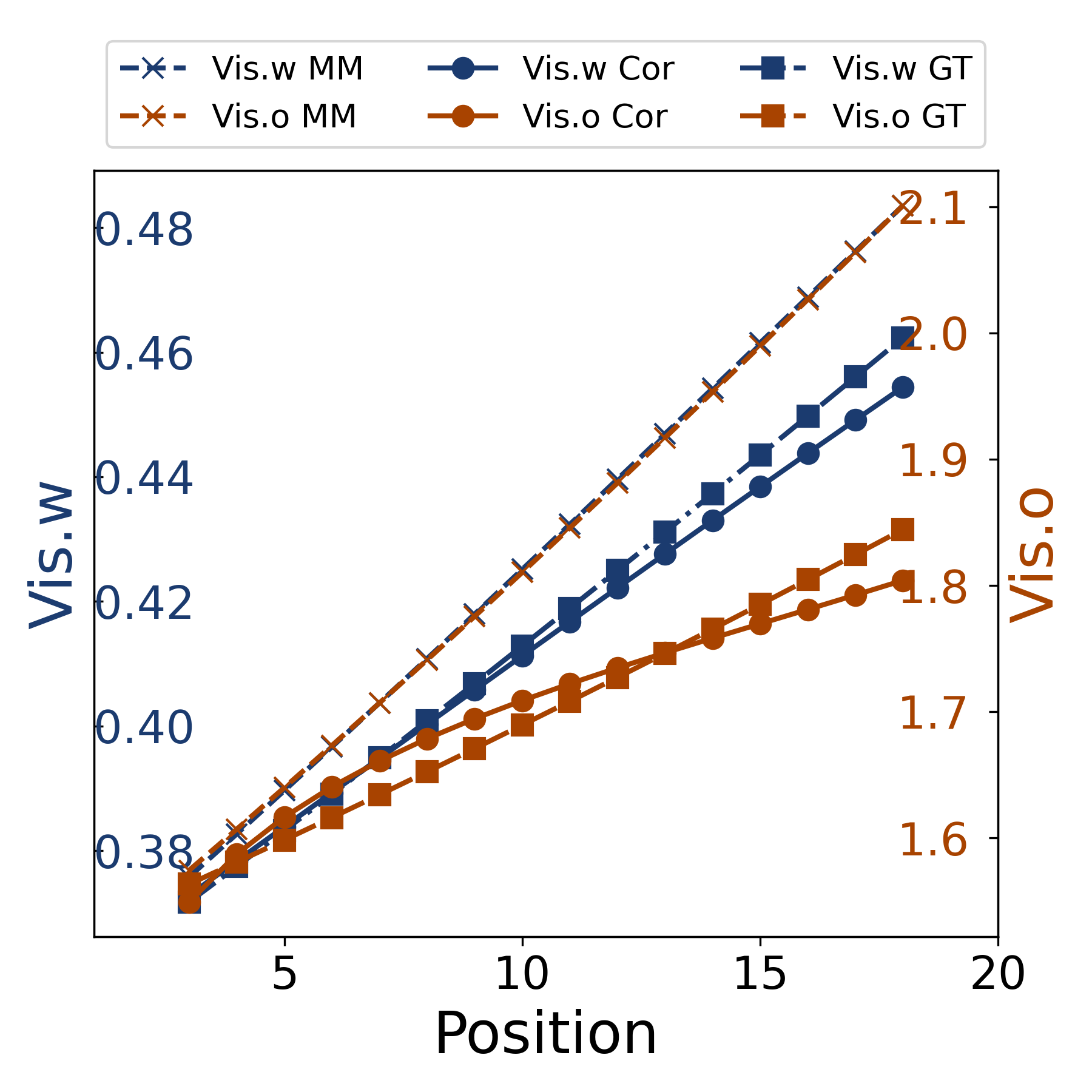}
    \label{fig12}
\end{minipage}
\hfill
\begin{minipage}[t]{0.3\textwidth}
    \centering
    \includegraphics[width=\textwidth]{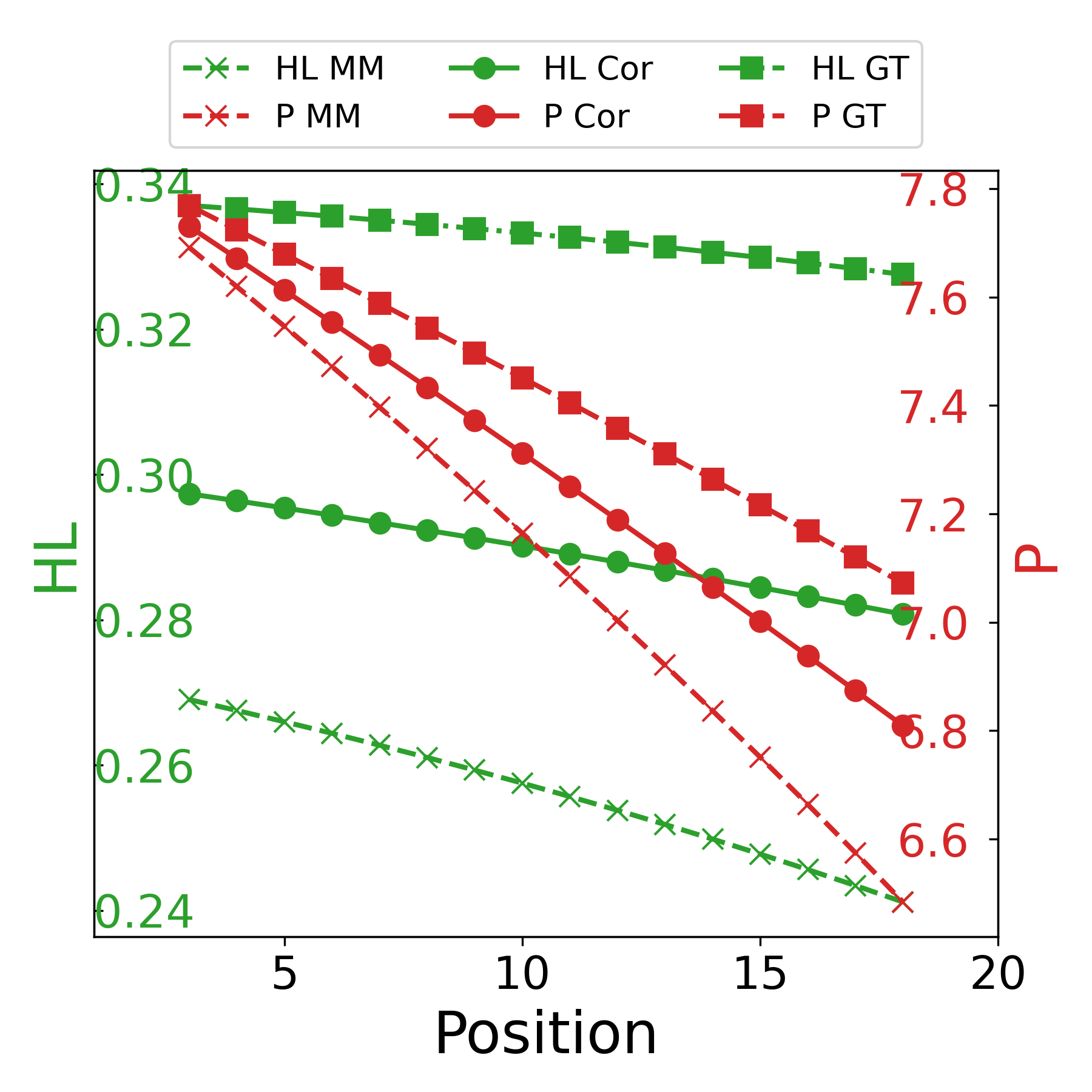}
    \label{fig13}
\end{minipage}
\hfill
\begin{minipage}[t]{0.3\textwidth}
    \centering
    \includegraphics[width=\textwidth]{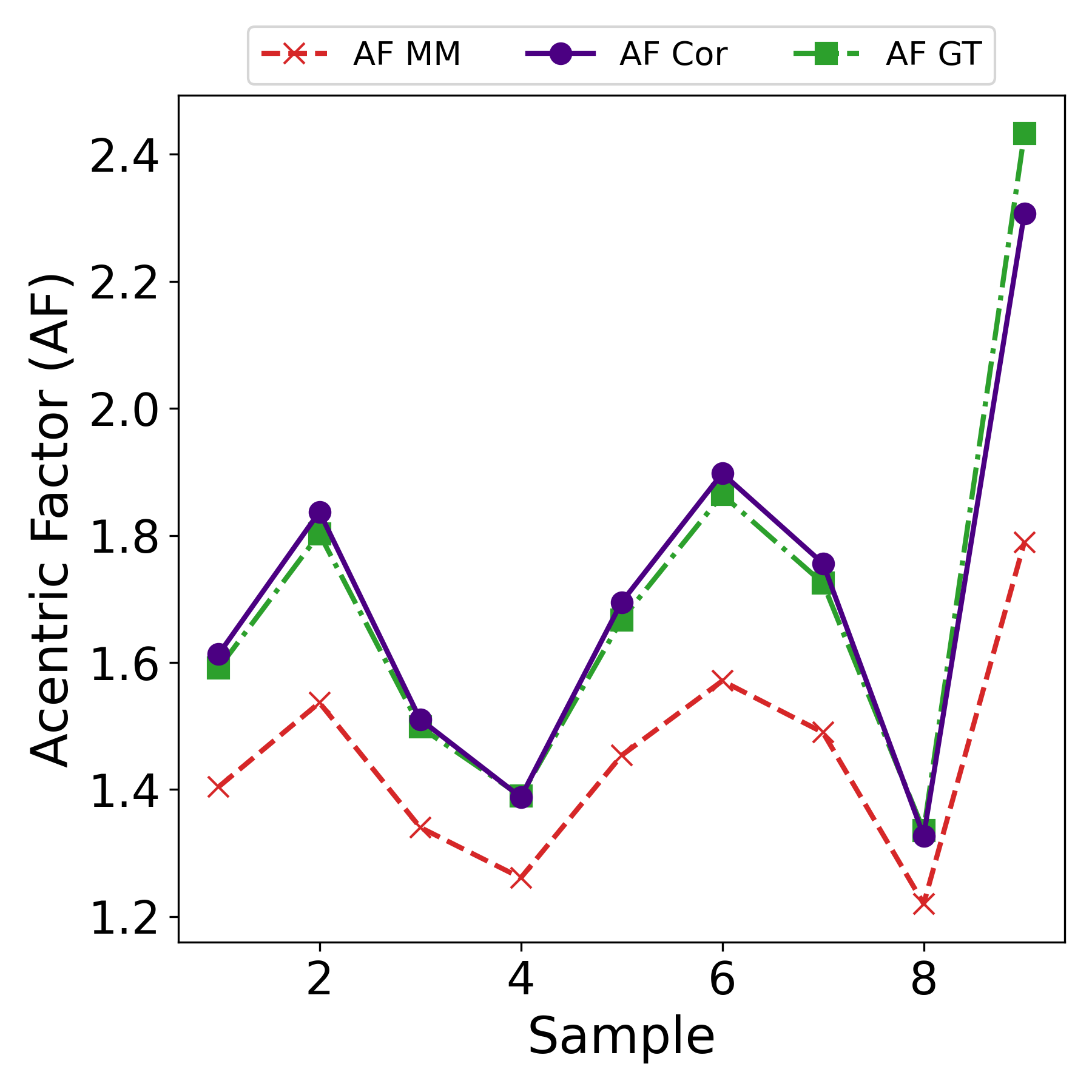}
    \label{fig14}
\end{minipage}

In summary, this work lays the groundwork for future advances in developing high-fidelity models for industrial digital twins, particularly from a symbiotic modeling perspective in which mechanistic models and data-driven models interact effectively, mutually influence each other, and jointly improve their respective performance. While we acknowledge that there remains a long way to go, DNA represents a pioneering effort that will be continuously refined, aiming to offer an effective pathway toward a new paradigm of next-generation symbiotic modeling.

\end{document}